\documentclass{article}

\usepackage[final,main]{neurips_2026}

\usepackage[utf8]{inputenc}
\usepackage[T1]{fontenc}
\usepackage{courier}
\usepackage{hyperref}
\usepackage{url}
\usepackage{booktabs}
\usepackage{amsfonts}
\usepackage{amsmath}
\usepackage{amssymb}
\usepackage{mathtools}
\usepackage{microtype}
\usepackage{xcolor}
\usepackage{graphicx}
\usepackage{subcaption}
\usepackage{multirow}
\usepackage{placeins}
\usepackage[capitalize,noabbrev]{cleveref}
\hypersetup{hidelinks,
  pdftitle={Geometric Alignment without Functional Equivalence: A Layer-wise Analysis of the Speech-Text Modality Gap},
  pdfauthor={Ming-Hao Hsu, Xiaohai Tian, Jun Zhang, Lu Lu, Yuxuan Wang, Zhizheng Wu}}

\title{Geometric Alignment without Functional Equivalence: A Layer-wise Analysis of the Speech--Text Modality Gap}

\author{Ming-Hao Hsu\textsuperscript{1} \enspace Xiaohai Tian\textsuperscript{2} \enspace Jun Zhang\textsuperscript{2} \enspace Lu Lu\textsuperscript{2} \enspace Yuxuan Wang\textsuperscript{2} \enspace Zhizheng Wu\textsuperscript{1,3,4}\thanks{\textsuperscript{1}School of Data Science, The Chinese University of Hong Kong, Shenzhen.
\textsuperscript{2}ByteDance.
\textsuperscript{3}City University of Macau.
\textsuperscript{4}Amphion Technology Co., Ltd.
Correspondence to: Zhizheng Wu \textless\href{mailto:wuzhizheng@cuhk.edu.cn}{\texttt{wuzhizheng@cuhk.edu.cn}}\textgreater.}}

\begin{document}

\maketitle

\begin{abstract}
End-to-end speech large language models (LLMs) often answer the same question less accurately from speech than from text, even when the spoken content matches the text prompt.
Existing analyses study this modality gap through the geometry of representations and measure how closely speech representations align with text.
Geometric alignment gives the gap a measurable form, but it does not show whether the model uses aligned representations in the same way.
This work connects representation geometry with answer formation and asks whether geometric alignment implies functional equivalence.
We compare cross-layer Centered Kernel Alignment (CKA) with layer-wise answer margins for four open-weight speech LLMs on SpeechMMLU and VoiceBench BBH.
In Qwen2.5-Omni and MiniCPM-o, speech and text representations align in middle layers while their late-layer answer margins still differ, so mid-layer alignment does not imply functional equivalence.
Qwen2-Audio instead shows weak alignment, which indicates a different failure mode.
An attention intervention at fixed input and token count identifies a causal effect of late attention allocation on answer-code separation, a quantity distinct from answer correctness.
Compressing the speech signal or its token sequence repairs a subset of the errors that occur only with speech.
\end{abstract}

\begin{figure}[h!]
    \centering
    \includegraphics[width=0.98\linewidth]{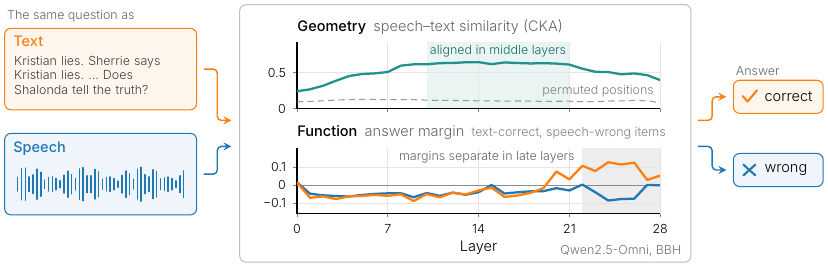}
    \caption{\textbf{Geometric alignment without functional equivalence.}
    The same question enters a speech LLM as text and as speech.
    Speech--text CKA is high in middle layers, yet the answer margins of the two inputs separate in late layers.
    Curves use Qwen2.5-Omni on BBH, with CKA from \cref{sec:appendix:native_cka} and margins averaged over the 233 items answered correctly only from text.}
    \label{fig:overview}
\end{figure}

\section{Introduction}
\suppressfloats[t]

\begin{figure}[t]
    \centering
    \includegraphics[width=0.98\linewidth]{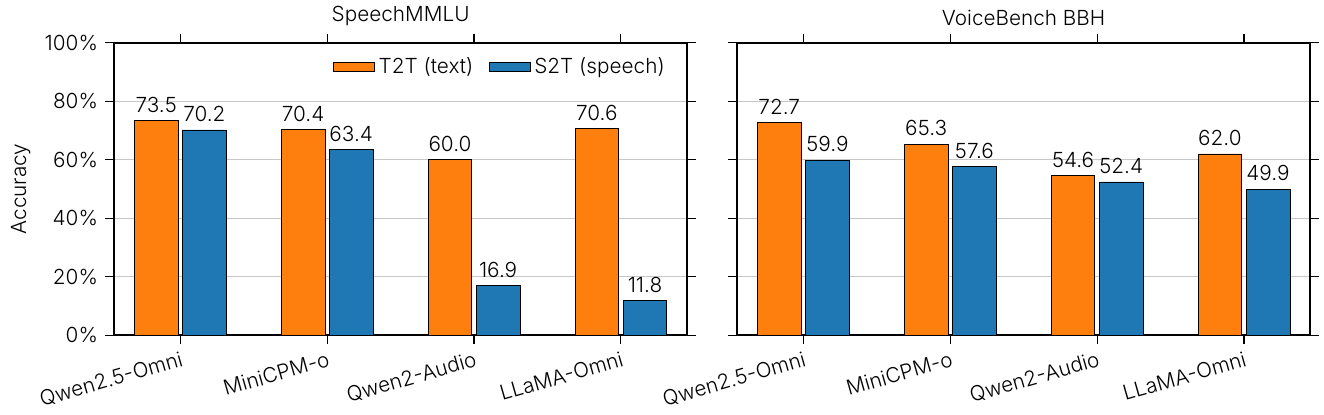}
    \caption{\textbf{The speech--text modality gap.}
    In the benchmark overview, speech-to-text (S2T) accuracy is below matched text-to-text (T2T) accuracy for all four speech LLMs on both SpeechMMLU and VoiceBench BBH.}
    \label{fig:backbone_acc_bars}
\end{figure}

Speech-based assistants aim to make the reasoning capabilities of language models available through spoken interaction~\citep{qwen_audio,speechgpt,audio_flamingo_3}.
End-to-end speech large language models (LLMs) pursue this goal by passing speech representations directly into an LLM, so that a spoken question can draw on the same knowledge and reasoning as a written one.
When an answer depends on linguistic content alone, the spoken and written versions of a question carry the same information, and the model should reach the same answer from either.
Yet speech LLMs answer such questions less accurately from speech than from text (\cref{fig:backbone_acc_bars}), a difference known as the modality gap~\citep{voicebench,s2sbench_2025,wang2026closing,hsu2026entity}.
The reasoning ability that the model already shows on text is therefore not fully available through speech.

A common way to study this gap is through the geometry of representations.
Speech and text reach the LLM through different input pathways, and prior work measures how closely the resulting representations align inside the model~\citep{alas,understandingthegap}, while related analyses describe modality-specific regions in vision-language models~\citep{mindthegap}.
This view gives the gap a measurable form and motivates aligning speech representations more closely with text.
Yet geometric similarity describes what representations look like, and it does not by itself show how the model uses them to reach an answer~\citep{ding2021grounding,semantics_at_an_angle,park2024linear}.

\suppressfloats[t]

A complementary line of work studies how a language model forms its answer across layers.
Layer-wise readouts such as the logit lens and the tuned lens show that a final answer is typically preceded by intermediate evidence formation and late-layer sharpening~\citep{logit_lens_ref,tuned_lens_ref,llm4stage}.
This view follows the answer directly.
For speech, however, layer-wise analyses have so far traced how acoustic features become semantic representations in recognition and classification systems~\citep{beyond_transcription_2025,behind_the_scenes_2025}.
It therefore remains unclear whether speech that has become geometrically aligned with text also goes on to form answers in the same way.

This work connects the two views to study where the modality gap arises inside the model (\cref{fig:overview}).
We ask whether geometric alignment implies functional equivalence, meaning that the model reaches the same answer from speech as from text.
Specifically, we present the same questions as speech and as text and measure where the two representations become similar across layers with Centered Kernel Alignment (CKA), a measure of similarity between two sets of representations.
At each layer, we also track how strongly the model favors the correct answer over competing ones.
We then edit attention inside the model while holding the input fixed, which tests whether late-layer answer formation depends causally on how attention is allocated over the speech input.

We study four open-weight speech LLMs, Qwen2.5-Omni~\citep{qwen2_5omni}, MiniCPM-o~\citep{minicpm_o}, Qwen2-Audio~\citep{qwen2_audio}, and LLaMA-Omni~\citep{llama_omni}, on SpeechMMLU~\citep{mmlu,mimoaudio} and VoiceBench BIG-Bench Hard (BBH)~\citep{voicebench,bbh}.
The results show that geometric alignment does not imply functional equivalence.
In Qwen2.5-Omni and MiniCPM-o, speech and text representations become aligned in middle layers, yet late layers still separate the correct answer from its competitors differently for the two inputs.
In Qwen2-Audio, the alignment itself is weak, so the gap takes different forms across models.
In Qwen2.5-Omni, the attention intervention shows that late-layer attention allocation causally affects how strongly the model separates candidate answers, irrespective of which answer is correct, and compressing the speech signal or its token sequence repairs a subset of speech-only errors.
These findings suggest that work on the modality gap should examine late-layer answer formation alongside representation alignment.

\section{Related Work}

\paragraph{Speech LLMs and the modality gap.}
Speech processing has moved from cascades of automatic speech recognition (ASR) and LLMs toward end-to-end speech LLMs~\citep{speechgpt,qwen_audio,llama_omni}.
Recent surveys organize this space into adapter-based models that map speech features into an LLM input space~\citep{qwen2_audio,minicpm_o,hsu2026cgate} and more unified systems that tokenize or model speech directly~\citep{audio_flamingo_3,when_llm_meet_speech_2025,recent_advances_slm_2025}.
These systems perform well on recognition, translation, and instruction following.
Yet speech-input reasoning still lags behind matched text-input reasoning~\citep{voicebench,s2sbench_2025,wang2026closing,hsu2026entity}, and work on closing this gap targets model-level score improvement~\citep{wang2026closing}.
In contrast, we study where the speech and text routes differ within a model, using paired questions whose answers are determined by linguistic content.

\paragraph{Geometric analyses of the modality gap.}
In vision-language models, \citet{mindthegap} show that different modalities occupy separate cones of a shared embedding space.
For speech, ALAS~\citep{alas} measures latent speech--text alignment layer by layer, and \citet{understandingthegap} analyze speech--text representational mismatch empirically.
Both motivate stronger feature alignment.
Yet analyses of cosine similarity~\citep{semantics_at_an_angle}, of linear representations~\citep{park2024linear}, and of how similarity measures track functional behavior~\citep{ding2021grounding} caution that geometric similarity alone does not determine how a model behaves.
We instead test whether speech that is geometrically aligned with text also supports the same late-layer answer formation.

\paragraph{Layer-wise analysis of answer formation.}
The logit lens~\citep{logit_lens_ref} and the tuned lens~\citep{tuned_lens_ref} decode intermediate hidden states into predictions, and stage-wise analyses show that final-layer answers are often preceded by intermediate evidence formation and late residual sharpening~\citep{llm4stage}.
For speech, mechanistic studies trace how acoustic features evolve into semantic representations in ASR or classification systems~\citep{beyond_transcription_2025,behind_the_scenes_2025}.
We bring this perspective to speech reasoning, comparing matched speech and text inputs with cross-layer CKA, decision-position attention, and answer-margin trajectories.

\paragraph{Speech redundancy and compression.}
Speech models face a density mismatch: speech token rates can be much higher than semantic word rates~\citep{entropy_based_speech_2025,hsu2025tasla}, and transformer speech stacks contain substantial redundancy~\citep{redundant_transformer_stack}.
Prior compression work often treats this redundancy mainly as an efficiency issue~\citep{liu2025speechtokenpredictioncompressedtofine}.
In contrast, we use compression as a diagnostic probe, comparing which speech-input errors temporal compaction and token merging repair (\cref{sec:speech_repair_probes}), and we examine attention allocation alongside answer separation.

\section{Experimental Setup and Diagnostics}

\paragraph{Task and behavioral gap.}
We compare the same semantic query under text-to-text (T2T) and speech-to-text (S2T) prompting.
For paired items $\mathcal D$, let $y_i$ be the reference answer and $\hat y_i^T,\hat y_i^S$ the parser-normalized text and speech answers.
The dataset-level modality gap is the difference in exact-match accuracy:
\begin{equation}
\label{eq:dataset_gap}
\Delta_{\mathcal D}=\frac{1}{|\mathcal D|}\sum_{i\in\mathcal D}
\left[\mathbf 1(\hat y_i^T=y_i)-\mathbf 1(\hat y_i^S=y_i)\right].
\end{equation}
A positive value means matched speech input underperforms matched text input.

\paragraph{Models, datasets, and scope.}
We study Qwen2.5-Omni-7B~\citep{qwen2_5omni}, MiniCPM-o 2.6~\citep{minicpm_o}, Qwen2-Audio-7B-Instruct~\citep{qwen2_audio}, and LLaMA-Omni~\citep{llama_omni}.
\Cref{fig:backbone_acc_bars} reports their T2T and S2T accuracies.
Layer 0 denotes the embedding output, and layers 1--28 denote successive Transformer-block outputs for Qwen2.5-Omni and MiniCPM-o.
SpeechMMLU~\citep{mmlu, mimoaudio} adapts MMLU to paired text and audio questions, while VoiceBench BBH~\citep{voicebench} adapts BIG-Bench Hard~\citep{bbh} to speech reasoning tasks.
Both benchmarks are semantics-matched: the answer is determined by linguistic content rather than speaker identity, emotion, prosody, or interaction structure.
We additionally use MMSU~\citep{mmsu2025} as a speech-native boundary check for Qwen2.5-Omni and MiniCPM-o.
MMSU includes tasks where stress, intonation, pauses, emotion, or background audio can be answer-relevant.
It tests the scope of conclusions drawn from semantics-matched questions.

\begin{figure*}[t]
    \centering
    \includegraphics[width=0.98\linewidth]{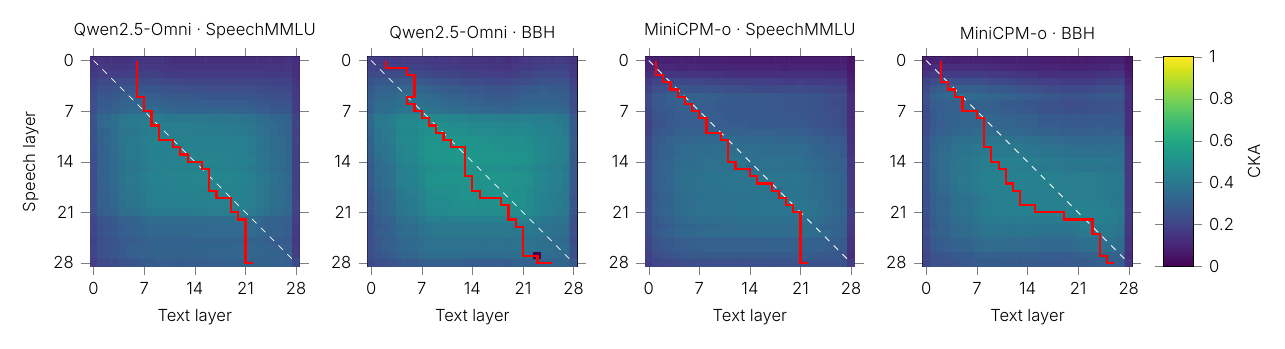}
    \caption{\textbf{Speech and text align in a broad middle-layer band.}
    Each heatmap shows cross-layer CKA between S2T layers on the y-axis and T2T layers on the x-axis after dynamic time warping (DTW) token alignment, with brighter colors denoting higher CKA.
    The red path marks the best-matching text layer for each speech layer, and the dashed diagonal marks equal layer indices.}
    \label{fig:cka_heatmap}
\end{figure*}

\paragraph{ASR-to-LLM baseline.}
To compare the direct speech route with a transcription interface, we evaluate an ASR-to-LLM pipeline that transcribes the audio and prompts the same downstream LLM with the transcript.
For multiple-choice settings, we preserve or restore explicit question and option markers when present, because answer formatting is part of the task interface.

\paragraph{Representation similarity.}
\label{sec:structural_alignment}
We compare speech and text hidden states after aligning their different sequence lengths.
Hidden states are extracted from the last user turn.
Let $H^S \in \mathbb{R}^{T_s \times D}$ and $H^T \in \mathbb{R}^{T_t \times D}$ be speech and text hidden states at a selected base layer.
We compute cosine similarities $M_{i,j}$ and use Dynamic Time Warping (DTW) to find a monotonic path from speech frames to text tokens.
Because each audio clip is paired with the same ordered text prompt, monotonic speech--text alignment is a reasonable diagnostic assumption for this setting.
We estimate this token correspondence once per item and freeze it across CKA layer pairs.
The main CKA analyses use base layer 10 for SpeechMMLU and 20 for BBH.
Both come from the stable region of a sweep over base layers 0, 5, 10, and 20, in which base layer 0 is the most sensitive (\cref{sec:appendix:dtw_sensitivity}).
When multiple speech frames align to the same text token, we mean-pool those frames, producing paired matrices $X,Y \in \mathbb{R}^{N \times D}$ for each example.
Our main similarity metric is linear CKA~\citep{kornblith2019similarity}, computed on the aligned speech matrix $X$ and text matrix $Y$:
\[
\mathrm{CKA}(X,Y)=\frac{\langle HKH, HLH\rangle_F}{\|HKH\|_F\|HLH\|_F}, \quad K=XX^\top,\ L=YY^\top.
\]
Here $H$ is the centering matrix and $N$ is the number of aligned token positions within the example.
Metrics are computed per example and averaged over the fixed evaluation subset (\cref{sec:appendix:setup_details}).
We also report standardized L2 distance and layer-wise mean token norm as controls for geometric proximity and activation scale.

\paragraph{Scope of the DTW alignment.}
The frozen path is a fixed measurement frame for the geometry analysis.
It relates residual-stream positions, without assuming that deep states contain only local information.
A path that is uninformative at depth would misalign rows and lower CKA toward the position-permutation baseline, so it cannot create the middle-layer band.
The best-matching-layer curve drawn on a CKA heatmap is a separate row-wise argmax, not the token-DTW path.
DTW is used only for geometry, and the margin readouts and attention edits use no cross-modal token alignment.

\paragraph{Answer formation.}
We characterize answer formation using layer-wise candidate margins.
The \emph{decision position} is the final valid prompt position after applying the chat template and its assistant-generation prefix.
Readouts at this position measure next-token candidate preference.
Let $z_{i,\ell}^c$ be the readout score for candidate answer $c$ at item $i$ and layer $\ell$.
For the paired BBH readout, intermediate states pass through the model's final RMS normalization (RMSNorm) and the shared vocabulary head, so hidden-state magnitude is divided out before logits are formed.
The final layer uses native logits without a second normalization.
Our primary margin is $m^p_{i,\ell}=q_{i,\ell}^{y_i}-\max_{c\ne y_i}q_{i,\ell}^c$, where $q=\operatorname{softmax}(z)$ over the task's answer candidates at temperature one.
Candidate probabilities express relative preference within the answer set and remain sensitive to logit scale.
We retain the logit margin $m^z_{i,\ell}=z_{i,\ell}^{y_i}-\max_{c\ne y_i}z_{i,\ell}^c$ as a complementary coordinate.
\Cref{sec:appendix:paired_readout} specifies candidate surface forms and the separate free-generation evaluation that assigns correctness groups.

\paragraph{Attention.}
We measure attention dispersion at the final prompt query with entropy, peak mass, and token coverage, and edit it at fixed input to test its effect on answer separation (\cref{sec:appendix:attention_intervention}).

\section{Empirical Analysis}
\label{sec:exp:phases}
\label{sec:empirical_analysis}

We first describe the behavioral gap and then follow the two views from the Introduction: where speech and text representations align, and where their answer formation diverges.
Unless otherwise stated, the detailed diagnostics use Qwen2.5-Omni, with additional model diagnostics in the Appendix.
The benchmark overview, 1,024-token diagnostic, and shared-instruction evaluation use different generation settings, summarized in \cref{tab:protocol_map}.

\subsection{The Behavioral Gap Is Task-Dependent}
\label{sec:behavioral_gap}

\Cref{fig:backbone_acc_bars} shows that S2T accuracy is below matched T2T accuracy for all four models, with a gap size that depends on the model and benchmark.
\Cref{tab:behavior_controls} details the gap for Qwen2.5-Omni.
On BBH, an ASR-to-LLM cascade improves over direct S2T by 7.5 percentage points but remains 5.3 points below gold-text T2T in the benchmark overview, while on SpeechMMLU the cascade is worse than direct S2T.
A Whisper-large-v2 audit of 200 items per benchmark gives a mean item word error rate (WER) of 4.65\% on BBH and 3.28\% on SpeechMMLU (\cref{tab:asr_wer_main}).
The cascade comparison combines transcription effects with changes in the downstream input interface, so the accuracy differences do not separate those effects causally.
The BBH categories are adjective ordering (Hyperbaton), navigation, sports plausibility, and truth-chain inference (Web-of-lies), which asks the model to follow truth relations over a chain of statements.
The gap is highly task-dependent, with Web-of-lies driving most of the aggregate difference~\citep{hsu2026entity}.
The comparison reverses when acoustic cues are relevant to the answer.
On the 2,420-item MMSU reasoning subset, deterministic rescoring gives an S2T advantage over text-only prompting of 34.3 points for Qwen2.5-Omni and 26.9 points for MiniCPM-o, with contributions from acoustic cues and answer-formatting behavior (\cref{tab:appendix:mmsu_scope_dense}).
The findings below therefore concern semantics-matched questions.

\begin{table}[t]
\centering
\caption{\textbf{Behavioral gap for Qwen2.5-Omni.}
Subtable (a) gives benchmark-overview accuracy with deltas in points relative to direct S2T, where SpMMLU denotes SpeechMMLU.
Subtable (b) gives the BBH categories Hyperbaton, Navigate, Sports, and Web-of-lies in the 1,024-token diagnostic, where the gap is T2T minus S2T.}
\label{tab:behavior_controls}
\vspace{1mm}
\small
\begin{minipage}[t]{0.53\linewidth}
\centering
\setlength{\tabcolsep}{4pt}
\renewcommand{\arraystretch}{1.05}
\textbf{(a) Behavior and ASR control}\par\vspace{0.5mm}
\begin{tabular}{@{}lrrrr@{}}
\toprule
\textbf{Setting} & \textbf{SpMMLU} & \textbf{$\Delta$} & \textbf{BBH} & \textbf{$\Delta$} \\
\midrule
Gold T2T & 73.49 & +3.25 & 72.7 & +12.8 \\
Direct S2T & 70.24 & base & 59.9 & base \\
ASR-to-LLM & 59.23 & \textminus11.0 & 67.4 & +7.5 \\
\bottomrule
\end{tabular}
\end{minipage}
\hfill
\begin{minipage}[t]{0.45\linewidth}
\centering
\setlength{\tabcolsep}{4pt}
\renewcommand{\arraystretch}{1.05}
\textbf{(b) BBH category gap}\par\vspace{0.5mm}
\begin{tabular}{@{}lcccc@{}}
\toprule
\textbf{Setting} & \textbf{Hyp.} & \textbf{Nav.} & \textbf{Sport} & \textbf{WoL} \\
\midrule
Gold-text T2T & 75.2 & 58.0 & 64.8 & 89.6 \\
Direct S2T & 72.0 & 58.0 & 56.8 & 52.0 \\
Gap & +3.2 & +0.0 & +8.0 & +37.6 \\
\bottomrule
\end{tabular}
\end{minipage}
\end{table}

\subsection{Speech and Text Align in Middle Layers}
\label{sec:geometry}

The cross-layer CKA heatmaps in \cref{fig:cka_heatmap} show three regimes along the speech layers, which we call Phases I to III.
In early layers, speech representations are not close to any text layer.
In middle layers, similarity rises and forms a broad band rather than a sharp diagonal, so several text layers have similarly high CKA with the same speech layer.
On SpeechMMLU, each speech layer aligns with roughly six text layers near its peak CKA value for both Qwen2.5-Omni and MiniCPM-o.
We use \emph{Semantic Smearing} as a descriptive name for this broad band and for the attention dispersion examined in \cref{sec:attention}.
In late layers, consecutive speech layers often share the same best-matching text layer (\cref{fig:phase_boundaries}).
The band persists without pooling.
When each text token is paired with one matched native speech frame, avoiding interpolation and temporal averaging, the middle-layer alignment lift and band width exceed their text-position-permutation baselines for both models on both benchmarks, with 95\% intervals above zero (\cref{tab:controls}a).
In a separate exploratory comparison, calibrating speech hidden states to match text means and standard deviations reduces BBH accuracy from a 62.3\% baseline to 46.8\% at the input layer and 61.2\% at the last layer.

\subsection{Answer Margins Diverge in Late Layers}
\label{sec:function}

\begin{figure}[t]
    \centering
    \includegraphics[width=0.98\linewidth]{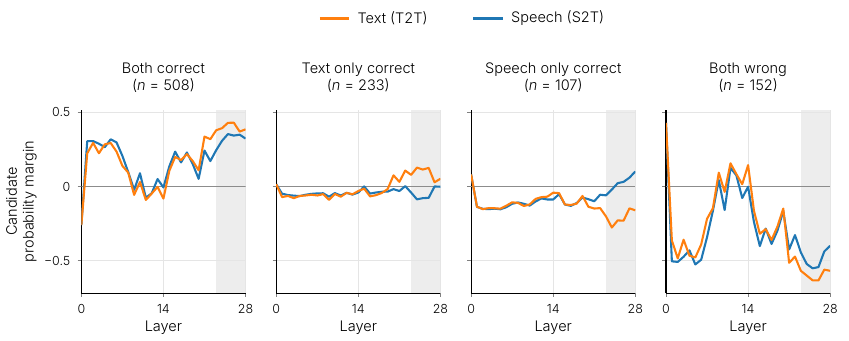}
    \caption{\textbf{Answer margins separate in late layers.}
    Curves show the mean correct-minus-competing candidate probability for Qwen2.5-Omni on 1,000 BBH items, grouped by free-generation correctness, and shading marks the late window of layers 23--28.
    In this window, the text margin is larger for text-only-correct items and the speech margin is larger for speech-only-correct items.}
    \label{fig:layerwise_margin_by_group}
\end{figure}

We next examine how the two routes separate the correct answer from competing candidates.
\Cref{fig:layerwise_margin_by_group} shows candidate-probability margins for the paired BBH readout, with items grouped by the correctness of free generation.
We compute a matched-instance late-margin difference over the final six layers, $\mathcal L_{\mathrm{late}}=\{23,\ldots,28\}$.
This fixed window covers roughly the last 20\% of the 28-layer backbone and lies within the CKA late-phase window of layers 22--28:
\begin{equation}
\label{eq:matched_late_margin_gap}
\delta_i^p=\frac{1}{|\mathcal L_{\mathrm{late}}|}\sum_{\ell\in\mathcal L_{\mathrm{late}}}
\left(m_{i,\ell}^{p,\mathrm{T2T}}-m_{i,\ell}^{p,\mathrm{S2T}}\right).
\end{equation}
The mean is +0.1351 for the 233 text-only-correct items and \textminus0.2301 for the 107 speech-only-correct items.
The difference between these group means, which we call the crossover contrast, is +0.3652 with a category-by-group paired-bootstrap 95\% interval of [0.2867,0.4426].
The contrast is also positive for full-vocabulary probability margins and signed logit margins (\cref{tab:controls}b).
In both groups, the route that answers correctly has the larger late-layer margin.
These comparisons are conditioned on generation outcomes, so they locate where the routes differ without establishing a cause.
Additional projected-logit and probe plots appear in \cref{sec:appendix:main_extra_diagnostics}.

\begin{table}[t]
\centering
\caption{\textbf{Controls for the two central findings.}
Subtable (a) gives the middle-layer CKA lift $A$ and band width $W_{.95}$ without pooling, each minus its position-permutation baseline, with 95\% bootstrap intervals (\cref{sec:appendix:native_cka}).
Subtable (b) gives the mean late T2T-minus-S2T margin of Qwen2.5-Omni on BBH for the two discordant groups in three margin coordinates, with the crossover contrast and its 95\% interval.}
\label{tab:controls}
\vspace{1mm}
\small
\setlength{\tabcolsep}{6pt}
\renewcommand{\arraystretch}{1.05}
\textbf{(a) Alignment without pooling}\par\vspace{0.5mm}
\begin{tabular}{@{}llcc@{}}
\toprule
\textbf{Model} & \textbf{Benchmark} & \textbf{$\Delta A$} & \textbf{$\Delta W_{.95}$} \\
\midrule
Qwen2.5-Omni & SpeechMMLU & .0952 [.0867, .1034] & 7.56 [6.88, 8.22] \\
Qwen2.5-Omni & BBH & .0981 [.0925, .1038] & 5.64 [5.07, 6.21] \\
MiniCPM-o & SpeechMMLU & .1343 [.1250, .1438] & 5.17 [4.52, 5.80] \\
MiniCPM-o & BBH & .0916 [.0840, .0993] & 5.47 [4.98, 5.95] \\
\bottomrule
\end{tabular}
\par\vspace{2.5mm}
\textbf{(b) Late-margin crossover}\par\vspace{0.5mm}
\begin{tabular}{@{}lrrc@{}}
\toprule
\textbf{Margin} & \textbf{Only T2T correct} & \textbf{Only S2T correct} & \textbf{Crossover [95\% CI]} \\
\midrule
Candidate probability & +.1351 & \textminus.2301 & +.3652 [.2867, .4426] \\
Full-vocabulary probability & +.0569 & \textminus.0527 & +.1096 [.0876, .1322] \\
Signed logit & +.8991 & \textminus.8906 & +1.7897 [1.4431, 2.1331] \\
\bottomrule
\end{tabular}
\end{table}

\subsection{Late Attention Allocation Causally Affects Answer Separation}
\label{sec:attention}

\begin{figure}[t]
    \centering
    \includegraphics[width=0.98\linewidth]{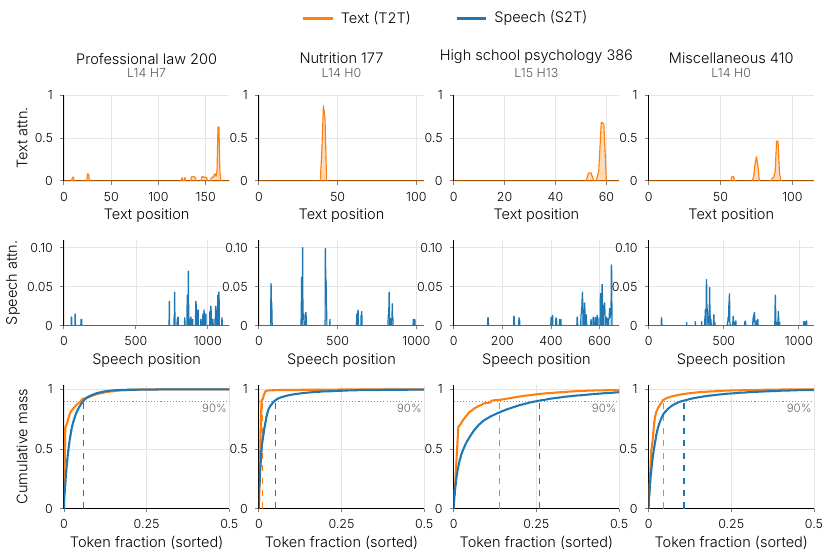}
    \caption{\textbf{Speech attention is more dispersed than text attention.}
    Each column shows a selected decision-position attention distribution for Qwen2.5-Omni on SpeechMMLU, labeled by subject, item, layer (L), and head (H).
    The speech row uses a ten-times smaller vertical scale, and dashed lines mark the sorted-token fraction that covers 90\% of the mass.}
    \label{fig:attention_dilution}
\end{figure}

Speech and text also differ in how attention is allocated at the decision position.
Across every head of Qwen2.5-Omni on 100 SpeechMMLU items, the median speech/text ratio of the tokens needed to cover 90\% of the attention is 2.20\texttimes{} in layers 7--21 and 3.74\texttimes{} in layers 23--28, and the speech-minus-text coverage difference is positive for 78.4\% and 86.1\% of head--layer pairs (\cref{tab:appendix:all_head_attention}).
\Cref{fig:attention_dilution} shows selected examples.
Dispersion is a route-level property: within speech, mean attention entropy provides no clear separation of correct and incorrect items (\cref{sec:appendix:attention_intervention}).

\begin{figure}[t]
    \centering
    \includegraphics[width=0.98\linewidth]{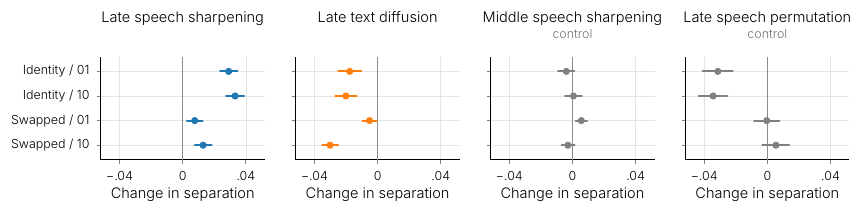}
    \caption{\textbf{Sharpening late speech attention increases answer-code separation.}
    Points show the mean change in $|z_X-z_Y|$ for Qwen2.5-Omni on 1,000 BBH items with 95\% bootstrap intervals, under two answer-to-code mappings and two rule orders.
    The input and token count are fixed, and the gap measures separation irrespective of which code is correct.}
    \label{fig:attention_intervention}
\end{figure}

Attention weights alone do not establish causal use~\citep{serrano2019attention}, so we test whether attention allocation affects answer separation with an intervention~\citep{vig2020investigating,shim2026linear} that edits attention while the input and token count stay fixed.
We encode each task's two answers as X and Y and counterbalance the answer-to-label mapping and the order of the two mapping instructions.
We then sharpen decision-position speech attention in layers 23--28 while preserving each head's total attention mass on the speech span (\cref{sec:appendix:attention_intervention}).
The change in the absolute two-choice logit gap, $\Delta|z_X-z_Y|$, measures how strongly either code is preferred, irrespective of correctness.
\Cref{fig:attention_intervention} shows that late speech sharpening increases this gap in all four conditions, while the reciprocal diffusion of text attention decreases it in all four.
Among the controls, equal-depth middle-layer sharpening has a 95\% interval entirely above zero in one condition, and location permutation in none.
The conditions reuse the same items, and baseline semantic predictions vary under answer-label swapping, failing the prespecified 75\% agreement criterion in three of four modality/order comparisons.
The causal interpretation therefore concerns answer-code separation.

\subsection{The Gap Takes Two Forms Across Models}
\label{sec:failure_modes}

The four-model comparison reveals different combinations of geometric alignment and answer behavior.
For Qwen2.5-Omni and MiniCPM-o, speech reaches a clear middle-layer alignment band with text (\cref{fig:cka_heatmap}), while the late answer margins depend on the route (\cref{fig:layerwise_margin_by_group,sec:appendix:minicpm}).
We call this \emph{Decision Instability}.
Qwen2-Audio instead has weak cross-modal CKA without a clear high-similarity middle band, a pattern we call \emph{Representation Failure}.
LLaMA-Omni, reported in the Appendix, shows broad similarity with less layer-specific correspondence.
\Cref{tab:failure_modes_main} summarizes these descriptive patterns, which suggest different targets for subsequent interventions: representation formation when alignment is weak and late answer separation when middle-layer alignment is present.

\begin{table}[t]
\centering
\caption{\textbf{Two failure modes across models.}
Gaps are T2T-minus-S2T accuracy points in the benchmark overview, approximate for LLaMA-Omni.
Mid-band describes the middle-layer CKA band.}
\label{tab:failure_modes_main}
\vspace{1mm}
\small
\setlength{\tabcolsep}{6pt}
\renewcommand{\arraystretch}{1.05}
\begin{tabular}{@{}lcccr@{}}
\toprule
\textbf{Model} & \textbf{SpeechMMLU gap} & \textbf{BBH gap} & \textbf{Mid-band} & \textbf{Mode} \\
\midrule
Qwen2.5-Omni & +3.25 & +12.8 & clear & Decision Instability \\
MiniCPM-o & +7.02 & +7.7 & clear & Decision Instability \\
LLaMA-Omni & $\approx$\,+58.8 & $\approx$\,+12.1 & broad & Intermediate \\
Qwen2-Audio & +43.07 & +2.2 & absent & Representation Failure \\
\bottomrule
\end{tabular}
\end{table}

\subsection{Compression Repairs a Subset of Speech-Only Errors}
\label{sec:speech_repair_probes}

We finally ask which speech errors compression can repair, comparing temporal compaction of the waveform with token merging inside the model on items that gold-text prompting answers correctly and direct speech answers incorrectly (\cref{tab:repair_and_merge_main}).
In the overview protocol, tempo 1.2\texttimes{} repairs 61 of 178 such items, including 50 of the 123 with ASR-correct transcripts, silence trimming repairs 68, and token merging repairs 26.
Within this protocol, temporal compaction repairs 2.35 to 2.62 times as many items as token merging.
In the shared-instruction protocol, the three interventions repair 113, 104, and 99 of 233 items, with paired difference intervals that include zero (\cref{sec:appendix:strict_behavior}).
Under both protocols, compaction and merging each repair a subset of these errors, while their relative advantage is protocol-dependent.
Token merging is an inference-only control on audio tokens after the early projection phase (\cref{sec:appendix:kv_merge}).
It changes the merged states as well as their count, so it does not isolate sequence length.
A text-side proxy that repeats each word also lowers T2T accuracy (\cref{tab:text_redundancy_causal}).
The comparisons identify distinct targets for testing repairs: cross-modal representation formation, allocation of attention within the speech span, and late answer separation.
Which intervention improves generated answers remains an empirical question for each model and task.

\begin{table}[t]
\centering
\caption{\textbf{Compression repairs a subset of speech-only errors on BBH.}
Gap repaired counts corrected answers among the text-correct, speech-wrong items of Qwen2.5-Omni under each protocol, and $\Delta$Acc. is the aggregate accuracy change in points.}
\label{tab:repair_and_merge_main}
\vspace{1mm}
\small
\setlength{\tabcolsep}{6pt}
\renewcommand{\arraystretch}{1.07}
\begin{tabular}{@{}llcccc@{}}
\toprule
& & \multicolumn{2}{c}{\textbf{Overview protocol}} & \multicolumn{2}{c}{\textbf{Shared-instruction protocol}} \\
\cmidrule(lr){3-4} \cmidrule(lr){5-6}
\textbf{Intervention} & \textbf{Target} & \textbf{Gap repaired} & \textbf{$\Delta$Acc.} & \textbf{Gap repaired} & \textbf{$\Delta$Acc.} \\
\midrule
Tempo 1.2\texttimes{} & audio duration & 61/178 (34.3\%) & +2.2 & 113/233 (48.5\%) & +2.2 \\
Silence trim & pauses & 68/178 (38.2\%) & +2.0 & 104/233 (44.6\%) & +1.3 \\
Token merging & speech tokens & 26/178 (14.6\%) & +0.5 & 99/233 (42.5\%) & \textminus1.5 \\
\bottomrule
\end{tabular}
\end{table}

\FloatBarrier
\section{Discussion}
\label{sec:discussion}

\paragraph{What the evidence establishes.}
The crossover contrast localizes natural outcome differences to late-layer answer margins, and the attention edit shows that late attention allocation changes answer-code separation at fixed input and token count.
These are different estimands, so their ratio is not a fraction of the gap explained by attention.
The edit establishes the sensitivity of answer-code separation, and it leaves accuracy repair, semantic direction, and other models untested.

\paragraph{Alternative explanations.}
Less reasoning-heavy speech training data could shape a weaker speech route and is compatible with the patterns we report.
Our analysis localizes where a fixed model diverges, whereas identifying the upstream cause requires checkpoints trained on matched data mixtures.
We therefore do not attribute the gap uniquely to architecture or to residual acoustic information.
Whether late speech states retain acoustic or phonetic variation that interferes with answer formation remains a hypothesis, and direct layer-wise feature probing is the test that would discriminate it.

\section{Conclusion and Limitations}
\label{sec:conclusion}

We study the speech--text modality gap by connecting two views of it: how similar speech and text representations are, and how the model forms its answer from them.
Geometric alignment does not imply functional equivalence.
In Qwen2.5-Omni and MiniCPM-o, speech representations become geometrically close to text in middle layers while still failing to form stable late-layer answer margins, whereas in Qwen2-Audio the alignment itself is weak.
A fixed-input attention edit in Qwen2.5-Omni shows that late attention allocation causally affects answer-code separation, a quantity distinct from answer correctness.
Temporal compaction and token merging each repair a subset of speech-only errors, with relative repair rates that vary across evaluation protocols.
These results suggest that evaluations of the modality gap should distinguish geometric alignment, late-layer answer margins, and generated-answer correctness, and that interventions should target representation formation when alignment is weak and late answer separation when alignment is present.

\paragraph{Limitations.}
The scope is semantics-matched reasoning over synthesized prompts and accessible open-weight speech LLMs around the 7B scale.
Speech-native tasks can require acoustic cues, as MMSU shows.
The phase description is model-dependent, and the intervention is limited to Qwen2.5-Omni on BBH.
Differences in training data may also contribute to the observed model differences.
The diagnostics do not identify which acoustic or phonetic features contribute to the speech errors.
Natural conversational speech, larger models, SpeechMMLU compaction, and post-compaction Phase-II/III diagnostics remain open tests.

\clearpage
\begin{ack}
The work of Zhizheng Wu is supported by the National Natural Science Foundation of China under Grant 62376237.
This work is supported by the Program for Guangdong Introducing Innovative and Entrepreneurial Teams, Grant No.\ 2023ZT10X044.
\end{ack}

{\small
\bibliographystyle{plainnat}
\bibliography{refs}
}

\clearpage
\appendix
\section*{Appendix}

The appendix follows the order of \cref{sec:empirical_analysis}.
\Cref{sec:appendix:robustness,sec:appendix:margins,sec:appendix:attention,sec:appendix:behavior} support the alignment, answer-margin, attention, and behavioral results.
\Cref{sec:appendix:minicpm,sec:appendix:llama,sec:appendix:qwen2audio} report the remaining models, and \cref{sec:appendix:qualitative} gives selected cases.
\Cref{sec:appendix:implementation,sec:appendix:broader_impact,sec:appendix:assets} give implementation details, broader impact, and asset licenses.

\FloatBarrier
\section{Robustness of the Alignment Analysis}
\label{sec:appendix:robustness}

We examine how the alignment diagnostics vary with the DTW base layer, distance metric, and band constraint, and whether the middle-layer pattern persists without pooling.

\subsection{DTW Sensitivity and Band Statistics}
\label{sec:appendix:dtw_sensitivity}

The cross-layer alignment analysis in \Cref{sec:structural_alignment} uses DTW to align speech-frame and text-token positions at a fixed base layer.
This token correspondence is held fixed when comparing different layer pairs.
A cross-layer best-match curve records the matched text-layer index at each speech layer.

\paragraph{Setup.}
The DTW base layer $\ell_0$ is swept over 0, 5, 10, and 20, with cosine versus $L_2$ cost computed after feature-wise standardization within each route, and with or without a width-64 Sakoe--Chiba band.
The sweep uses Qwen2.5-Omni with BF16 weights on 100 items per benchmark.
Let $p_\ell$ denote the matched text-layer index at speech layer $\ell$, for $\ell=0,\ldots,L-1$.
We report the $R^2$ of a linear fit $p_\ell=a\ell+b$ and the stall fraction $\frac{1}{L-1}\sum_{\ell=1}^{L-1}\mathbf{1}[p_\ell=p_{\ell-1}]$, which counts adjacent speech-layer pairs mapped to the same text layer.
The fit measures linearity rather than distance to the identity diagonal.

\begin{figure}[htbp]
    \centering
    \includegraphics[width=0.98\linewidth]{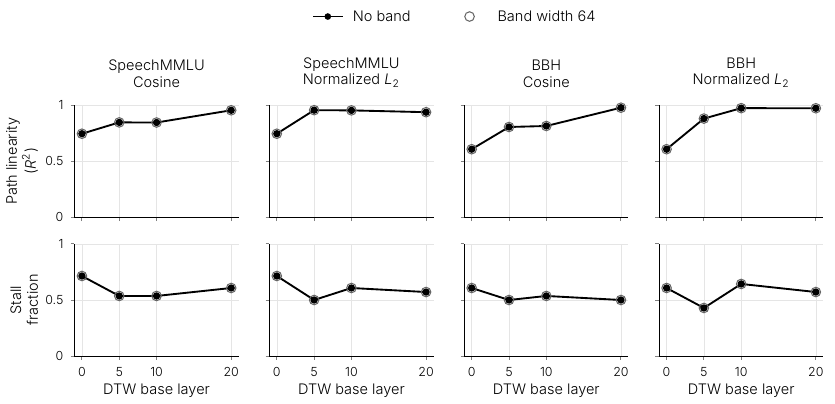}
    \caption{\textbf{Cross-layer best-match sensitivity to DTW settings.}
Qwen2.5-Omni, 100 items per benchmark.
The top row reports the $R^2$ of a linear fit to matched text-layer indices across speech layers, and the bottom row the fraction of adjacent speech-layer pairs mapped to the same text layer.
Open rings mark the width-64 band and coincide with the unbanded values.
Pooling-free, null-adjusted results appear in \cref{sec:appendix:native_cka}.}
    \label{fig:dtw_sensitivity_r2_stall}
\end{figure}

\paragraph{Results.}
The sweep in \cref{fig:dtw_sensitivity_r2_stall} shows greater sensitivity at base layer 0 than at the higher bases.
Path linearity is $R^2\le0.75$ at base layer 0 against $R^2\ge0.80$ at every higher base in all four panels, and the stall fraction is highest at base layer 0 in three of the four panels.
The width-64 band leaves both statistics unchanged in every setting.
We retain bases 10 and 20 for SpeechMMLU and BBH, respectively.
The native-length, position-permutation controls in \cref{sec:appendix:native_cka} supply the quantitative test of whether the middle-layer lift and band survive without pooling.
All layer-pair comparisons use the same estimated token correspondence.

\subsection{Native-Length Alignment without Interpolation or Averaging}
\label{sec:appendix:native_cka}

We test whether the Phase-II CKA lift and broad near-peak band persist without the interpolation and temporal averaging used in the original heatmaps.
We freeze 100 items per benchmark and use the same IDs for Qwen2.5-Omni and MiniCPM-o, yielding four model--benchmark cells with no failed items.
Both models use BF16 weights without quantization, processing one item at a time.
The BBH sample contains 25 items from each of its four categories, and the SpeechMMLU sample is balanced across subjects.
For each item, we isolate the final user content span, excluding padding, instructions outside that span, and assistant positions.
We compute one cosine-distance DTW path on native hidden states at layer 10 for SpeechMMLU or layer 20 for BBH and reuse it at every layer pair.
For each text token, we select the lower median of the unique speech-frame indices assigned by that path.
Thus each CKA row contains one native speech state and one native text state, with no interpolation, truncation, or within-group averaging.
Linear CKA uses float64 centered-Gram arithmetic after conversion of the hidden states to float32.

For each item, let $A$ be the Phase-II mean of the row-wise CKA maxima minus the corresponding Phase-I mean.
Let $W_{.95}$ be the Phase-II mean number of text layers reaching at least 95\% of each speech layer's maximum CKA.
We retain the original phase windows: 0--9, 10--21, and 22--28 for Qwen2.5-Omni on both benchmarks, 0--3, 4--20, and 21--28 for MiniCPM-o on SpeechMMLU, and 0--12, 13--23, and 24--28 for MiniCPM-o on BBH.
We compare each endpoint with its within-item mean over 20 fixed non-identity text-position permutations, reusing each permutation at every layer pair.
We bootstrap paired items 10,000 times with seed 20260724, stratifying BBH by category, and report percentile 95\% intervals.
Items are the resampling unit, with permutations and layer pairs summarized within each item.

\begin{table}[htbp]
\centering
\small
\setlength{\tabcolsep}{4pt}
\caption{\textbf{Native-length CKA controls.}
The first two endpoints subtract the within-item mean position-permutation endpoint and report item means with 95\% bootstrap intervals.
The last column gives the alignment-free matched-minus-wrong-item differences in $A$ and $W_{.95}$, respectively, without an inferential interval.}
\label{tab:appendix:native_cka}
\begin{tabular}{@{}llccc@{}}
\toprule
\textbf{Model} & \textbf{Benchmark} & \textbf{$\Delta A$} & \textbf{$\Delta W_{.95}$} & \textbf{Cross-item $(\Delta A,\Delta W_{.95})$} \\
\midrule
Qwen2.5-Omni & SpeechMMLU & .0952 [.0867, .1034] & 7.56 [6.88, 8.22] & (.0299, \phantom{\textminus}1.37) \\
Qwen2.5-Omni & BBH & .0981 [.0925, .1038] & 5.64 [5.07, 6.21] & (.0081, \textminus0.72) \\
MiniCPM-o & SpeechMMLU & .1343 [.1250, .1438] & 5.17 [4.52, 5.80] & (.0521, \textminus0.46) \\
MiniCPM-o & BBH & .0916 [.0840, .0993] & 5.47 [4.98, 5.95] & (\textminus.0031, \textminus3.30) \\
\bottomrule
\end{tabular}
\end{table}

In all four model--benchmark combinations (\cref{tab:appendix:native_cka}), the CKA lift and band width exceed their position-permutation means, with positive 95\% intervals.
The lower confidence bound for the observed $W_{.95}$ also exceeds one.
We separately remove token alignment by computing CKA across the 100 paired final-content states, one state per item and layer.
We compare matched items with 20 wrong-item derangements within category or subject.
Text layer 0 is undefined because the fixed final text token has zero cross-item variance, so matched and deranged matrices both use speech layers 0--28 and text layers 1--28.
Only Qwen2.5-Omni on SpeechMMLU satisfies all three directional criteria of positive lift above the derangement mean, width above one, and width above the derangement mean.
This analysis changes the CKA sample axis from token positions to items and provides mixed corroboration of alignment-free geometry.
The native-length result therefore supports independence from interpolation and mean pooling while retaining the assumption of a monotonic token correspondence.

We also compare mean pooling, center-frame selection, a random frame averaged over 20 selection seeds, the full repeated DTW path, and repeated group means after resampling both sequences to 256 positions.
The no-averaging conditions retain the lift and broad-band criteria in all four cells, including width thresholds of 90\% and 99\%.
These fixed-length controls isolate aggregation and path weighting, whereas the native-length control above additionally removes resampling.

\subsection{Best-Match Paths and Phase Windows}
\label{sec:appendix:phase_paths}

\begin{figure}[htbp]
    \centering
    \includegraphics[width=0.98\linewidth]{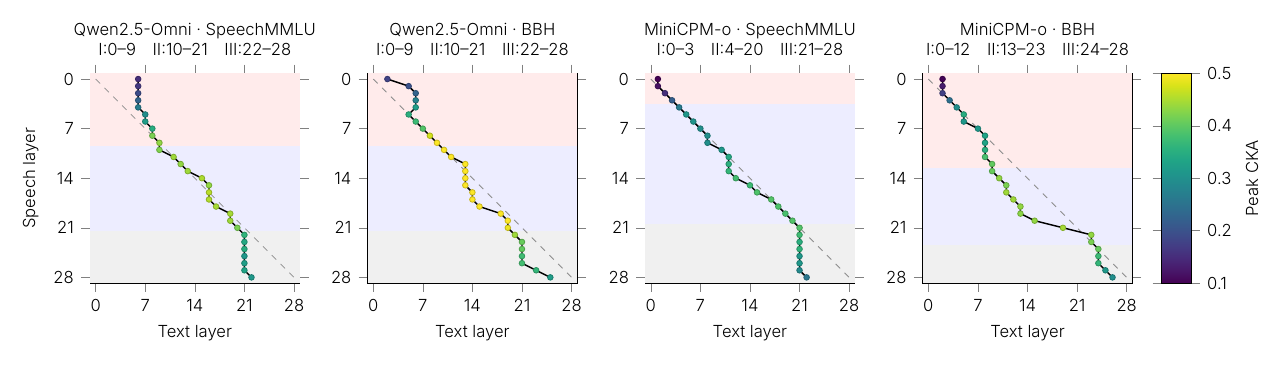}
    \caption{\textbf{Best-match paths and phase windows from the CKA heatmaps.}
    Black lines connect each speech layer to its best-matching text layer, point colors give the peak CKA in that speech-layer row, and the dashed diagonal marks equal layer indices.
Background shading marks the three phase windows, whose speech-layer ranges appear above each panel and are listed in \cref{sec:appendix:native_cka}.
The late plateau describes repeated best-matching text layers, with functional behavior tested separately by answer readouts.}
    \label{fig:phase_boundaries}
\end{figure}

\Cref{fig:phase_boundaries} redraws the main CKA heatmaps as best-match paths together with the three phase windows.

\FloatBarrier
\section{Answer-Margin Readouts}
\label{sec:appendix:margins}

\subsection{Paired Decision Readouts and Margin Definitions}
\label{sec:appendix:paired_readout}

We use all 1,000 BBH items and define correctness groups from the shared-instruction free-generation evaluation in \cref{sec:appendix:strict_behavior}.
The group sizes are 508 for both routes correct, 233 for only T2T correct, 107 for only S2T correct, and 152 for both routes wrong.
For the paired readout, we append two newline characters, \texttt{The answer is}, and a trailing space to each route's category instruction inside the user message.
We then apply the chat template with its assistant generation prefix and read the final valid prompt position.
Thus the probe predicts the next generated token from the assistant-prefix state, rather than continuing an assistant-side answer phrase.
This controlled prompt provides a common decision position and is distinct from the prompt used to assign free-generation correctness groups.
We retain all 29 hidden-state coordinates, comprising the embedding output and 28 Transformer outputs.
At intermediate coordinates we apply the model's final RMSNorm once, followed by the complete vocabulary projection.
At the final coordinate we index the model's native full-vocabulary logits directly, avoiding both a second normalization and rounding changes from a candidate-only projection.
Candidate scores are promoted to float32 before subtraction and softmax.

Hyperbaton uses the candidates A and B, with bare and leading-space uppercase forms.
The remaining categories use yes and no, each with lowercase and initial-capital forms, with and without a leading space.
We retain only single-token forms, and no token belongs to both candidates.
For candidate $c$, we define $z_c=\max_{v\in V_c}z_v$, where $V_c$ contains its retained surface tokens.
The signed logit margin is $z_y-\max_{c\ne y}z_c$, and the primary plotted probability margin is $q_y-\max_{c\ne y}q_c$, where $q=\operatorname{softmax}(z)$ at temperature one over the task's two candidates.
We also compute a full-vocabulary probability margin by summing the full-vocabulary softmax mass over $V_c$ for each candidate before taking the gold-minus-best-alternative difference.
These probability views have explicit denominators but remain sensitive to logit scale and are not calibrated confidence estimates.

\begin{figure}[htbp]
    \centering
    \includegraphics[width=0.98\linewidth]{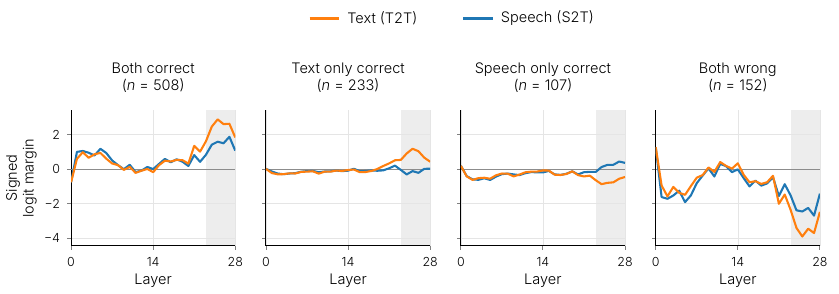}
    \caption{\textbf{Signed-logit counterpart of the probability replication.}
    The items, readout and groups match \cref{fig:layerwise_margin_by_group}.
    Curves show mean correct-minus-competing logits, with the fixed late window shaded.}
    \label{fig:appendix:signed_margin}
\end{figure}

For each margin, we average the paired T2T-minus-S2T difference over layers 23--28 within each item.
The crossover contrast subtracts the mean difference in the only-S2T-correct group from that in the only-T2T-correct group.
We bootstrap route-paired items independently within each category-by-correctness cell 10,000 times with seed 20260725, preserving the observed cell sizes.
\Cref{tab:controls} reports item-weighted group means and percentile intervals.
The candidate-probability crossover is positive in all four categories, whereas the full-vocabulary crossover is slightly negative for web of lies (\textminus0.00024).
Grouping on observed correctness makes this a localization of an outcome-conditioned difference, not a causal estimate or an independent correctness test.
\Cref{fig:appendix:signed_margin} gives the signed-logit counterpart of the paired probability analysis.

\subsection{Projected-Logit and Probe Diagnostics}
\label{sec:appendix:main_extra_diagnostics}

\begin{figure}[htbp]
    \centering
    \includegraphics[width=0.98\linewidth]{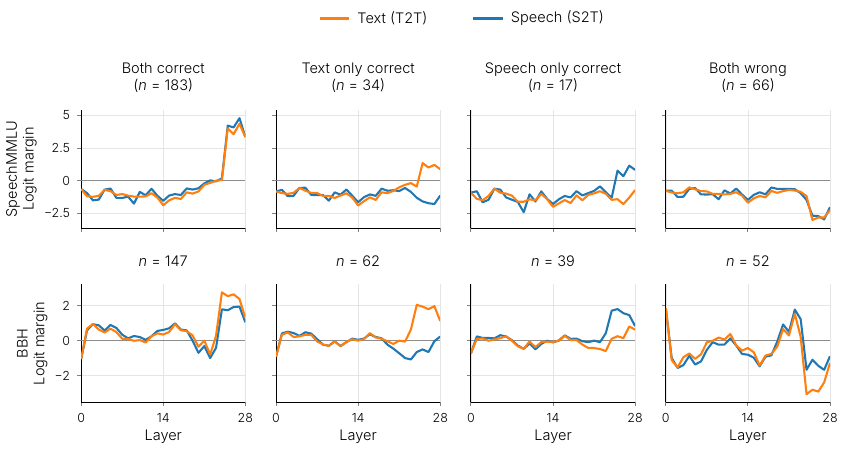}
    \caption{\textbf{Qwen2.5-Omni logit margin by correctness group.}
    Rows are SpeechMMLU and BBH, and columns are the free-generation groups with their sizes.
    Curves show the mean candidate-readout margin of the group.
    The BBH groups are those of the 1,000-item analysis in \cref{fig:layerwise_margin_by_group} restricted to 300 items.
    In the text-only-correct group the late-layer margin (mean of the last six coordinates) is +0.63 for text and \textminus1.40 for speech on SpeechMMLU, and +1.58 and \textminus0.46 on BBH.}
    \label{fig:appendix:submitted_logit_margin}
\end{figure}

\begin{figure}[htbp]
    \centering
    \includegraphics[width=0.98\linewidth]{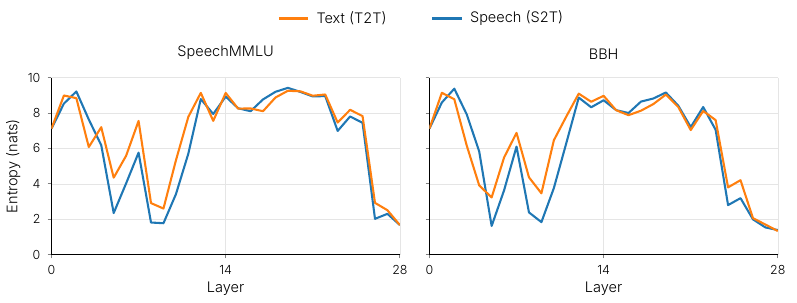}
    \caption{\textbf{Layer-wise projected-logit entropy of Qwen2.5-Omni.}
    Curves show the mean over 300 items per benchmark.
    Speech and text have similar entropy trajectories, with mean absolute differences of 0.67 nats on SpeechMMLU and 0.76 nats on BBH.}
    \label{fig:layerwise_entropy}
\end{figure}

\begin{figure}[htbp]
    \centering
    \includegraphics[width=0.7\linewidth]{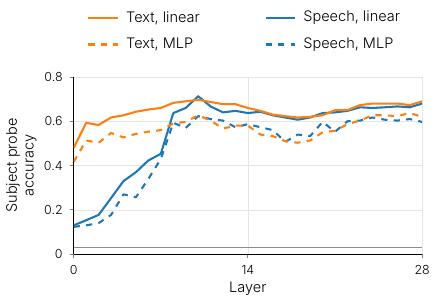}
    \caption{\textbf{Layer-wise subject probe of Qwen2.5-Omni.}
    Logistic-regression (linear) and one-hidden-layer MLP probes on the mean hidden state of the evidence tokens predict the SpeechMMLU subject of the 300 items (34 subjects, chance 2.9\%), with 5-fold cross-validated accuracy.
    Linear speech accuracy rises from 0.13 at the embedding output to 0.71 at layer 10 and text accuracy from 0.48 to 0.70, and both stay between 0.6 and 0.7 afterwards; fold standard deviations reach 0.10.
    These curves describe probe performance rather than generated-answer correctness.}
    \label{fig:layer_probe_curve}
\end{figure}

\begin{figure}[htbp]
    \centering
    \includegraphics[width=0.98\linewidth]{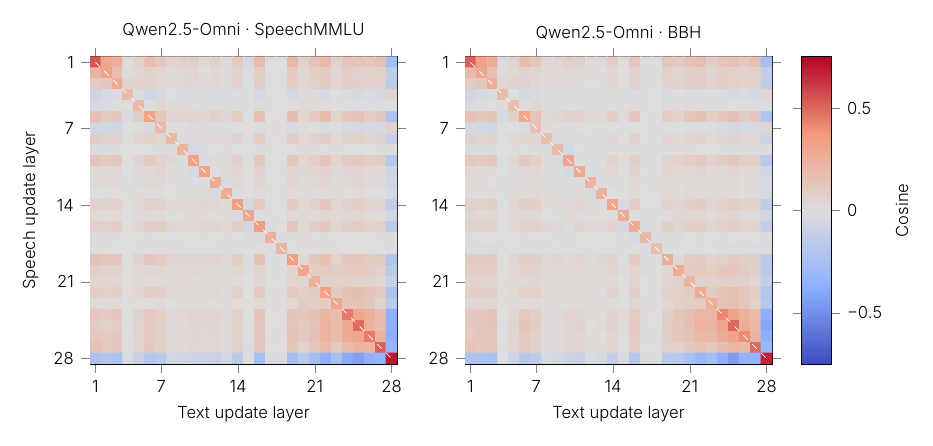}
    \caption{\textbf{Cosine similarity between speech and text layer-wise updates in Qwen2.5-Omni.}
    An update is the difference between consecutive hidden states after the fixed base-layer-0 token alignment, averaged over 100 items per benchmark.
    For every speech layer the largest cosine lies at the text layer with the same index.
    The mean diagonal cosine is 0.33 on SpeechMMLU and 0.30 on BBH, against 0.03 off the diagonal.}
    \label{fig:update_cosine}
\end{figure}

The entropy, candidate-readout and margin plots in this subsection and in \cref{sec:appendix:minicpm,sec:appendix:llama,sec:appendix:qwen2audio} read the final prompt position, apply the model's final RMSNorm once at intermediate coordinates, and index the native logits at the last coordinate, as in the paired analysis of \cref{sec:appendix:paired_readout}.
They use different prompts, the 300 aggregate items per benchmark of the figure reruns, and BF16 weights, so their absolute scales are not directly comparable with the paired readout.
Margins are candidate logits after an assistant answer cue, the correct option minus the best other option.
\Cref{fig:appendix:submitted_logit_margin} shows the margins for Qwen2.5-Omni, and \cref{fig:layerwise_entropy,fig:layer_probe_curve,fig:update_cosine} show the entropy, probe, and update diagnostics, respectively.

\paragraph{Correctness groups.}
For each model, benchmark and route, we decode the 300 items greedily with at most 256 new tokens, truncate at the first conversation-role marker, and parse the answer with the deterministic parser of \cref{sec:appendix:strict_behavior}, counting unparsed answers as wrong.
BBH uses the category instruction of that section, and SpeechMMLU a zero-shot instruction that ends with \texttt{The answer is (A).} to \texttt{(D).}; the text route receives the question followed by the same instruction.
Qwen2.5-Omni on BBH keeps the groups of the 1,000-item analysis, restricted to these items.
The groups depend on the free-generation answer alone, so they can disagree with the sign of a readout margin.
\Cref{tab:appendix:group_protocol} lists accuracies, unparsed answers and group sizes.
MiniCPM-o, Qwen2-Audio on BBH, and the speech route of LLaMA-Omni break the answer format more often, so their groups partly reflect formatting.

\begin{table}[htbp]
\centering
\small
\setlength{\tabcolsep}{4pt}
\caption{\textbf{Free-generation correctness groups of the 300 items per benchmark.}
Accuracy counts unparsed answers as wrong.
Group sizes are both correct, text only correct, speech only correct, and both wrong.
Qwen2.5-Omni on BBH uses the groups of \cref{sec:appendix:strict_behavior}, whose unparsed counts are reported there.}
\label{tab:appendix:group_protocol}
\begin{tabular}{@{}llrrcc@{}}
\toprule
\textbf{Model} & \textbf{Benchmark} & \textbf{S2T acc.} & \textbf{T2T acc.} & \textbf{Unparsed S2T / T2T} & \textbf{Group sizes} \\
\midrule
Qwen2.5-Omni & SpeechMMLU & 66.7 & 72.3 & 0 / 4 & 183 / 34 / 17 / 66 \\
Qwen2.5-Omni & BBH & 62.0 & 69.7 & \textendash{} & 147 / 62 / 39 / 52 \\
MiniCPM-o & SpeechMMLU & 50.0 & 57.3 & 64 / 73 & 101 / 71 / 49 / 79 \\
MiniCPM-o & BBH & 56.7 & 50.0 & 51 / 98 & 96 / 54 / 74 / 76 \\
Qwen2-Audio & SpeechMMLU & 33.3 & 55.3 & 10 / 8 & 70 / 96 / 30 / 104 \\
Qwen2-Audio & BBH & 49.0 & 42.3 & 26 / 81 & 78 / 49 / 69 / 104 \\
LLaMA-Omni & SpeechMMLU & 16.3 & 67.0 & 114 / 16 & 33 / 168 / 16 / 83 \\
LLaMA-Omni & BBH & 29.3 & 57.3 & 114 / 54 & 50 / 122 / 38 / 90 \\
\bottomrule
\end{tabular}
\end{table}

\FloatBarrier
\section{Attention Analyses}
\label{sec:appendix:attention}

\subsection{All-Head Attention Coverage}
\label{sec:appendix:all_head}

\Cref{tab:appendix:all_head_attention} gives the all-head attention coverage summary cited in \cref{sec:attention}, together with normalized entropy.

\begin{table}[htbp]
\centering
\caption{\textbf{All-head attention coverage audit for Qwen2.5-Omni on SpeechMMLU.}
We summarize all head--layer pairs within each item, then report the across-item medians of the within-item medians and 90th percentiles (Q90).
The positive-pair fraction is averaged over items.
$\Delta$ denotes S2T minus T2T, and Ent. is attention entropy divided by the log of valid prompt length.
Attention is normalized over all valid prompt keys at the final prompt query.
These windows differ from the CKA phase windows.
Speech usually needs more tokens to cover 90\% of the attention mass, while normalized entropy differences vary across heads.}
\label{tab:appendix:all_head_attention}
\vspace{1mm}
\small
\setlength{\tabcolsep}{5pt}
\renewcommand{\arraystretch}{1.06}
\begin{tabular}{@{}lrrrrrr@{}}
\toprule
\textbf{Layers} & \textbf{Items} & \textbf{Median cov. ratio} & \textbf{Median $\Delta$Cov\textsubscript{.90}} & \textbf{Q90 $\Delta$Cov\textsubscript{.90}} & \textbf{Pairs $\Delta$Cov $>$ 0} & \textbf{Q90 $\Delta$Ent.} \\
\midrule
7--21 & 100 & 2.20\texttimes{} & +16.0 & +83.5 & 78.4\% & +0.037 \\
23--28 & 100 & 3.74\texttimes{} & +48.0 & +102.5 & 86.1\% & +0.073 \\
\bottomrule
\end{tabular}
\end{table}

\subsection{Selected-Head Attention Summary}
\label{sec:appendix:selected_head}

\Cref{tab:attention_dilution_metrics} reports a selected-head attention diagnostic, with examples in \cref{fig:attention_dilution}.
The speech peak mass is 0.105 versus 0.638 for text, top-10 mass is 0.483 versus 0.933, and 90\% coverage requires 101 versus 8 tokens.
These selected summaries emphasize large cross-modal differences, and coverage counts depend on input length.

\begin{table}[htbp]
\centering
\caption{\textbf{Selected-head attention summary for Qwen2.5-Omni on SpeechMMLU.}
$\Delta$ is S2T minus T2T, computed from the displayed values.
Entropy\textsubscript{norm} is normalized attention entropy, $p_{\max}$ is the largest attention coefficient, top-10 mass sums the ten largest coefficients, and Cov\textsubscript{.90} counts the tokens needed to cover 90\% of attention.}
\label{tab:attention_dilution_metrics}
\vspace{1mm}
\small
\setlength{\tabcolsep}{6pt}
\renewcommand{\arraystretch}{1.05}
\begin{tabular}{@{}lccc@{}}
\toprule
\textbf{Metric} & \textbf{T2T} & \textbf{S2T} & \textbf{$\Delta$} \\
\midrule
Entropy\textsubscript{norm} & 0.361 & 0.657 & +0.296 \\
$p_{\max}$ & 0.638 & 0.105 & \textminus0.533 \\
Top-10 mass & 0.933 & 0.483 & \textminus0.450 \\
Cov\textsubscript{.90} count & 8 & 101 & +93 \\
\bottomrule
\end{tabular}
\end{table}

\subsection{Fixed-Input Attention Interventions and Their Interpretation}
\label{sec:appendix:attention_intervention}

We intervene on the final prompt query of every head in Qwen2.5-Omni, using evaluation mode, eager attention, and batch size one.
Within each baseline--treatment pair, the input tensors, token count, positions, and model weights are identical.
For the target content span, we normalize the post-softmax attention coefficients to a conditional distribution $p_j$, replace it by $p'_j=p_j^{1/\tau}/\sum_k p_k^{1/\tau}$, and restore the original per-head span mass.
All coefficients outside the span and the current value vectors remain unchanged by the local edit.
We apply the resulting attention-output difference and let later computation proceed normally, without matching the attention-output norm.
Speech sharpening uses $\tau$~=~.6 at layers 23--28.
We compare it with sharpening at the six middle layers 12--17, a within-span permutation of the sharpened late coefficients using seed 1234, and reciprocal text diffusion at layers 23--28 with $\tau$~=~1.6666667.
The permutation is a location control whose intervention magnitude is not matched to sharpening.

We cross each condition with two answer-label mappings and two rule orders on the same 1,000 items.
Semantic class 0 denotes A for hyperbaton and no otherwise, and class 1 denotes B or yes.
The identity mapping assigns classes 0 and 1 to X and Y, and the swapped mapping reverses this assignment.
Rule order states either the class-0 or the class-1 rule first, followed by \texttt{Output ONLY a single character: X or Y.}
For this intervention assay, each candidate score is the log-sum-exp of the native logits for its bare and leading-space forms, unlike the maximum-over-forms readout in \cref{sec:appendix:paired_readout}.
The prespecified direction-free diagnostic is the per-item change in $|z_X-z_Y|$ within each mapping/order cell.
We report the 16 cell estimates in \cref{tab:appendix:attention_cells}, using category-stratified paired-item bootstrap intervals with 10,000 resamples and seed 20260724.
The four cells reuse the same items and are not independent replications, and the intervals are marginal rather than simultaneous.

\begin{table}[htbp]
\centering
\small
\setlength{\tabcolsep}{4pt}
\caption{\textbf{Change in absolute answer-code separation.}
Each entry is the mean treatment-minus-baseline change with its 95\% interval on the same 1,000 items.
Positive values indicate greater separation regardless of which semantic answer is favored.
Order 01 states the class-0 rule first, and order 10 states the class-1 rule first.}
\label{tab:appendix:attention_cells}
\begin{tabular}{@{}llcc@{}}
\toprule
\textbf{Condition} & \textbf{Order} & \textbf{Identity mapping} & \textbf{Swapped mapping} \\
\midrule
S2T late sharpening & 01 & +.0291 [.0237, .0346] & +.0076 [.0026, .0127] \\
                    & 10 & +.0332 [.0274, .0389] & +.0129 [.0076, .0184] \\
S2T middle sharpening & 01 & \textminus.0039 [\textminus.0091, .0012] & +.0056 [.0019, .0093] \\
                      & 10 & +.0006 [\textminus.0048, .0059] & \textminus.0029 [\textminus.0069, .0013] \\
S2T late permutation & 01 & \textminus.0314 [\textminus.0410, \textminus.0219] & \textminus.0004 [\textminus.0084, .0077] \\
                     & 10 & \textminus.0344 [\textminus.0436, \textminus.0253] & +.0054 [\textminus.0032, .0138] \\
T2T late diffusion & 01 & \textminus.0177 [\textminus.0251, \textminus.0103] & \textminus.0051 [\textminus.0097, \textminus.0006] \\
                   & 10 & \textminus.0201 [\textminus.0268, \textminus.0133] & \textminus.0301 [\textminus.0351, \textminus.0250] \\
\bottomrule
\end{tabular}
\end{table}

Late speech sharpening increases separation in all four cells, whereas middle sharpening has an interval entirely above zero in one cell and the permutation control in none.
Reciprocal text diffusion decreases separation in all four cells.
Our criterion for interpreting a change as semantic improvement requires at least 75\% agreement between baseline semantic predictions under identity and swapped mappings for each modality and rule order.
Agreement is 73.7\% and 59.9\% for S2T, and 84.9\% and 69.7\% for T2T, so three of the four comparisons fall below the criterion.
The prespecified mapping-averaged correct-log-odds change is +.00381 [.00117, .00652], but label-dependent semantic predictions make this change ambiguous as a measure of answer improvement.
The intervention therefore establishes a change in absolute answer-code separation at fixed input.
The largest measured span-mass error is 5.96\texttimes10\textsuperscript{\textminus8}, the outside-span coefficient change is zero, and the $\tau$~=~1 check reproduces native logits exactly on its four-item check set.

We separately test whether matched text-query guidance moves the full speech next-token distribution toward the matched text distribution.
At the same late layers, we score current speech keys with the matched text query, mix the resulting speech-conditional attention distribution with the current one at weight .25, and preserve current values and total span mass.
On all 1,000 items, the reduction in full-vocabulary Jensen--Shannon divergence is +.000033 nats [\textminus.000202, .000273].
Its advantage over a category- and semantic-label-matched wrong-item query is +.000208 [\textminus.000015, .000438].
Both intervals include zero, leaving uncertain whether text-query guidance brings the speech output distribution closer to the text route or improves on wrong-item guidance.

\paragraph{Within-speech correctness comparison.}
On the 750 BBH items held out from the 250-item intervention pilot, we average normalized speech-span entropy across heads and layers 23--28 at the final prompt query.
The assay's baseline semantic readout gives 439 correct and 311 incorrect items, with mean entropies .73205 and .72795.
The equally category-weighted correct-minus-incorrect contrast is .00109 [\textminus.00032, .00251], using 10,000 category-by-correctness bootstrap resamples with seed 20260725.
The interval spans zero, providing no clear difference in mean entropy between correct and incorrect items under this readout.
The comparison uses the intervention assay's baseline readout, not the separate free-generation correctness groups.

\FloatBarrier
\section{Behavioral Protocols and Repair}
\label{sec:appendix:behavior}

\subsection{Behavioral Protocols}
\label{sec:appendix:protocols}

\Cref{tab:protocol_map} distinguishes the behavioral settings used in the paper.
The category breakdown and 178-item repair subset use a 1,024-token diagnostic evaluation, with archived scoring that includes fallback parsing.
The shared-instruction comparison instead uses a 256-token limit and deterministic parsing, with unparsed answers counted as wrong.
Its 233-item subset, paired readouts, and complete transition counts therefore form a separate comparison.

\begin{table}[htbp]
\centering
\small
\caption{\textbf{Behavioral settings and their baseline accuracies.}
All values are percentages for Qwen2.5-Omni on BBH.
A dash denotes a route not used for that comparison.
Baseline differences identify separate settings, so their accuracy changes and correctness groups are interpreted within each row.}
\label{tab:protocol_map}
\setlength{\tabcolsep}{4pt}
\begin{tabular*}{\linewidth}{@{\extracolsep{\fill}}lrrp{0.40\linewidth}@{}}
\toprule
Setting & T2T & S2T & Use in the paper \\
\midrule
Benchmark overview & 72.7 & 59.9 & Aggregate model comparison and overview repair changes \\
1,024-token diagnostic & 71.9 & 59.7 & Category breakdown, duration groups, 178-item gap subset \\
Calibration & --- & 62.3 & Input- and last-layer moment matching \\
Word repetition & 62.4 & --- & $r$~=~1 reference for the text stress test \\
Shared instructions, 256 tokens & 74.1 & 61.5 & 233-item gap subset, paired readouts, complete behavioral transitions \\
\bottomrule
\end{tabular*}
\end{table}

\subsection{Speech-Token Merging Control}
\label{sec:appendix:kv_merge}

The token-merging intervention in \cref{tab:repair_and_merge_main} tests compression after speech projection.
It is an inference-time mid-layer token-merging intervention applied only to the speech path of Qwen2.5-Omni on BBH, with no finetuning and no changes to the text path.
Compression starts after the early projection phase, where speech representations are already closer to text representations, and is applied across zero-based Thinker modules 7--21, corresponding to paper layers 8--22.

For each selected layer, we isolate the audio-token span, split audio hidden states into an even-index destination set and an odd-index source set, L2-normalize both sets, and assign each source token to its highest-cosine destination token.
The highest-scoring source--destination pairs are merged by weighted averaging until the configured removal budget is reached, after which the source tokens are removed and the attention mask and KV cache are rebuilt for the shortened sequence.
Non-audio tokens are never merged.
The control uses target keep ratio 0.5, minimum 64 audio tokens, maximum 512 audio tokens, zero cosine threshold, and a back-loaded merge schedule over these modules.
The intervention changes retained state content as well as token count, so its effects combine compression and representation changes.

\subsection{Shared-Instruction Evaluation and Complete Transitions}
\label{sec:appendix:strict_behavior}

This evaluation applies shared instructions to all behavioral conditions and reports both repaired and newly introduced errors.
This protocol yields 233 text-correct/speech-wrong items, compared with 178 in the 1,024-token diagnostic setting (\cref{tab:protocol_map}).
We evaluate Qwen2.5-Omni-7B on the same 1,000 VoiceBench BBH IDs, with 250 items in each of hyperbaton, navigate, sports understanding, and web of lies.
We use the official general system prompt and the identical modality-neutral category instruction for speech, gold text, and ASR transcripts.
Hyperbaton is instructed to end with \texttt{The answer is (A).} or \texttt{The answer is (B).}, and the remaining categories use the corresponding yes/no endings.
All routes allow reasoning and use greedy decoding with at most 256 new tokens.
Before the common deterministic parser, we truncate generated text at the first new conversation-role marker: \texttt{\textbackslash nHuman:}, \texttt{\textbackslash nUser:}, \texttt{\textbackslash nAssistant:}, \texttt{\textbackslash nsystem\textbackslash n}, \texttt{\textbackslash nuser\textbackslash n}, or \texttt{\textbackslash nassistant\textbackslash n}.
This format guard applies no additional output-token cap or prefix restriction.
An unparsed answer counts as wrong and remains in the denominator.
No item is skipped and no inference failure is recorded in any of the six conditions.
Gold-text accuracy is 74.1\% and direct-speech accuracy is 61.5\%, with 3 and 19 unparsed answers, respectively.
The runs use BF16 model weights without quantization, PyTorch 2.6.0 with CUDA 12.4, and Transformers 4.52.1.

We generate the tempo condition with SoX 14.4.2, using the pitch-preserving \texttt{tempo -s 1.2} effect.
Silence trimming uses 20-ms RMS windows with 10-ms hops, a threshold 40 dB below peak RMS, and 50-ms padding around retained regions.
Both transformations operate on 48-kHz mono PCM-32 audio, with stereo sources converted by channel averaging.
Silence trimming uses no normalization, crossfade, or resampling.
The aggregate retained waveform-frame fractions are .8333 for tempo and .8283 for silence, and neither transformation introduces additional clipped samples.
The merging condition follows \cref{sec:appendix:kv_merge} and records an aggregate full-sequence retention fraction of .5612.

\begin{table}[htbp]
\centering
\small
\setlength{\tabcolsep}{3pt}
\caption{\textbf{Complete paired transitions relative to direct speech.}
Repairs are wrong-to-correct transitions, harms are correct-to-wrong transitions, and retained C/W denotes unchanged correct/wrong outcomes across all 1,000 items.
Gap repairs use the 233 items correct only under gold text as their denominator.
Accuracy changes are percentage points with category-stratified paired-bootstrap 95\% intervals.}
\label{tab:appendix:strict_transitions}
\begin{tabular}{@{}lrrrrrrc@{}}
\toprule
\textbf{Condition} & \textbf{\shortstack{Acc.\\(\%)}} & \textbf{Unparsed} & \textbf{Repairs} & \textbf{Harms} & \textbf{\shortstack{Retained\\C/W}} & \textbf{\shortstack{Gap\\repairs}} & \textbf{$\Delta$Acc. [95\% CI]} \\
\midrule
Tempo 1.2\texttimes{} & 63.7 & 20 & 151 & 129 & 486/234 & 113/233 & +2.2 [\textminus1.1, 5.4] \\
Silence trimming & 62.8 & 22 & 139 & 126 & 489/246 & 104/233 & +1.3 [\textminus1.9, 4.5] \\
Token merging & 60.0 & 21 & 139 & 154 & 461/246 & 99/233 & \textminus1.5 [\textminus4.8, 1.8] \\
ASR cascade & 71.5 & 14 & 219 & 119 & 496/166 & 181/233 & +10.0 [6.7, 13.4] \\
\bottomrule
\end{tabular}
\end{table}

We use 10,000 category-stratified paired bootstrap resamples with base seed 20260725 and contrast-specific deterministic seeds, together with exact two-sided McNemar tests on discordant pairs.
The gap repair rates are 48.50\% [42.06, 54.94] for tempo, 44.64\% [38.20, 51.07] for silence, and 42.49\% [36.48, 48.50] for merging.
The paired tempo-minus-merging repair-rate difference is +6.01 points [\textminus2.15, 14.59], and the silence-minus-merging difference is +2.15 points [\textminus6.44, 10.31].
Thus the point estimates favor waveform compaction, while neither comparison establishes a repair advantage and the net-accuracy intervals for all three interventions include zero.

The cascade transcribes the original audio with Whisper large-v2 in English at temperature zero and supplies that transcript to the same text route.
We apply Unicode NFC and whitespace normalization without restoring gold-text content.
Across all 1,000 items, micro-averaged WER is 5.27\%, mean item WER is 5.03\%, and mean option-marker recall is 99.2\%.
The cascade gains 10.0 accuracy points over direct speech, but its difference from gold text is \textminus2.6 points [\textminus5.4, .1], with exact McNemar $p$~=~.0783.
The comparison establishes a benefit of the cascade over direct speech in this protocol while leaving the residual cascade--text difference uncertain.
The archived records include post-execution hashes of the inputs, transformed audio, transcripts, predictions, model files, code, and launch settings.
These identify the evaluated artifacts but do not guarantee bitwise reproducibility across kernels or hardware.

\subsection{Transcript Audit}
\label{sec:appendix:asr_audit}

\Cref{tab:asr_wer_main} reports the Whisper-large-v2 transcript audit behind the word error rates in \cref{sec:behavioral_gap}.
Format-marker recall is 99.5\% on BBH and 98.4\% on SpeechMMLU.

\begin{table}[htbp]
\centering
\caption{\textbf{ASR transcript audit.}
We transcribe 200 randomly sampled items per benchmark with Whisper-large-v2.
Word error rate (WER), substitution (Sub.), insertion (Ins.), and deletion (Del.) rates are averaged over items.
Med. WER is the median item WER.
Marker recall measures recognition of ``the answer is'' on BBH and that phrase plus ``question'' and ``option A/B/C/D'' on SpeechMMLU.}
\label{tab:asr_wer_main}
\vspace{1mm}
\small
\setlength{\tabcolsep}{6pt}
\renewcommand{\arraystretch}{1.05}
\begin{tabular}{@{}lrrrrrrr@{}}
\toprule
\textbf{Dataset / slice} & \textbf{N} & \textbf{WER} & \textbf{Med. WER} & \textbf{Marker} & \textbf{Sub.} & \textbf{Ins.} & \textbf{Del.} \\
\midrule
SpeechMMLU & 200 & 3.28\% & 0.00\% & 98.4\% & 1.89\% & 0.66\% & 0.72\% \\
BBH overall & 200 & 4.65\% & 3.45\% & 99.5\% & 4.09\% & 0.26\% & 0.29\% \\
\quad Hyperbaton & 45 & 2.55\% & 0.00\% & 100.0\% & 1.82\% & 0.60\% & 0.13\% \\
\quad Navigate & 51 & 3.09\% & 0.00\% & 100.0\% & 2.76\% & 0.06\% & 0.28\% \\
\quad Sports & 54 & 3.22\% & 2.44\% & 100.0\% & 2.48\% & 0.31\% & 0.43\% \\
\quad Web-of-lies & 50 & 9.66\% & 9.84\% & 98.0\% & 9.24\% & 0.11\% & 0.31\% \\
\bottomrule
\end{tabular}
\end{table}

\subsection{Audio Duration}
\label{sec:appendix:duration}

In the 1,024-token diagnostic, text-correct/speech-wrong items have longer audio than both-correct items on average, 23.58 versus 22.41 seconds, with a 95\% bootstrap interval of [0.44, 1.89] seconds for the difference (\cref{tab:appendix:bbh_temporal_load}).
This duration comparison is pooled across categories.

\begin{table}[htbp]
\centering
\caption{\textbf{BBH temporal-load metadata by correctness group.}
Durations use the 1,000-item, 1,024-token diagnostic setting in \cref{tab:protocol_map}.
Confidence intervals are 95\% bootstrap intervals over items, pooled across categories.}
\label{tab:appendix:bbh_temporal_load}
\vspace{1mm}
\small
\setlength{\tabcolsep}{6pt}
\renewcommand{\arraystretch}{1.06}
\begin{tabular}{@{}lccc@{}}
\toprule
\textbf{Group} & \textbf{Count} & \textbf{Duration (s)} & \textbf{Seconds per word} \\
\midrule
Both correct & 541 & 22.41 [22.04, 22.77] & 0.44 [0.43, 0.44] \\
Only T2T correct & 178 & 23.58 [22.97, 24.18] & 0.45 [0.44, 0.46] \\
Only S2T correct & 56 & 21.23 [20.32, 22.17] & 0.45 [0.43, 0.46] \\
Both wrong & 225 & 21.54 [20.97, 22.11] & 0.42 [0.42, 0.43] \\
\bottomrule
\end{tabular}
\end{table}

\subsection{Text-Redundancy Stress Test}
\label{sec:appendix:text_redundancy}

\Cref{tab:text_redundancy_causal} reports a text-side proxy stress test.
Repeating each word reduces T2T accuracy relative to the unrepeated input in both benchmark-specific stress tests.
The repeated text is out of distribution and changes both length and linguistic form.

\begin{table}[htbp]
\centering
\caption{\textbf{Text-redundancy stress test for Qwen2.5-Omni.}
Each word of the text prompt is repeated $r$ times, and entries are T2T accuracy in percent.}
\label{tab:text_redundancy_causal}
\vspace{1mm}
\small
\setlength{\tabcolsep}{6pt}
\renewcommand{\arraystretch}{1.05}
\begin{tabular}{@{}ccc@{}}
\toprule
\textbf{$r$} & \textbf{SpeechMMLU} & \textbf{BBH} \\
\midrule
1 & 73.9 & 62.4 \\
2 & 67.4 & 56.2 \\
4 & 57.3 & 55.4 \\
8 & 51.2 & 53.5 \\
\bottomrule
\end{tabular}
\end{table}

\subsection{Speech-Native Boundary on MMSU}
\label{sec:appendix:mmsu}

\Cref{tab:appendix:mmsu_scope_dense} reports selected speech-native MMSU tasks.
For MMSU, Qwen2.5-Omni has 1 versus 257 unparsed speech/text answers among 2,420 items, and MiniCPM-o has 1 versus 4.
Excluding seven shared invalid or ambiguous source labels preserves the speech advantage, so the boundary result combines acoustic cues with answer-formatting behavior.

\begin{table*}[htbp]
\centering
\caption{\textbf{MMSU results on selected speech-native tasks.}
Acoustic cues can be answer-relevant on MMSU.
Scores use deterministic parsing, counting unparsed answers as wrong.
These selected tasks have large speech advantages, with $\Delta$ denoting S2T minus text accuracy in percentage points.}
\label{tab:appendix:mmsu_scope_dense}
\vspace{1mm}
\small
\setlength{\tabcolsep}{6pt}
\renewcommand{\arraystretch}{1.05}
\begin{tabular}{@{}lrrrrrrr@{}}
\toprule
\textbf{Task} & \textbf{N} & \multicolumn{3}{c}{\textbf{Qwen2.5-Omni}} & \multicolumn{3}{c}{\textbf{MiniCPM-o}} \\
\cmidrule(lr){3-5} \cmidrule(lr){6-8}
& & \textbf{S2T} & \textbf{Text} & \textbf{$\Delta$} & \textbf{S2T} & \textbf{Text} & \textbf{$\Delta$} \\
\midrule
Continuation writing & 110 & 100.0 & 22.7 & +77.3 & 99.1 & 22.7 & +76.4 \\
Homophone reasoning & 101 & 97.0 & 26.7 & +70.3 & 95.0 & 35.6 & +59.4 \\
Intent detection & 115 & 88.7 & 4.3 & +84.3 & 87.0 & 25.2 & +61.7 \\
Polysemy reasoning & 98 & 99.0 & 43.9 & +55.1 & 96.9 & 38.8 & +58.2 \\
Deixis resolution & 103 & 94.2 & 24.3 & +69.9 & 83.5 & 51.5 & +32.0 \\
Dialogue reasoning & 114 & 83.3 & 32.5 & +50.9 & 86.8 & 41.2 & +45.6 \\
Code-switch QA & 111 & 90.1 & 44.1 & +45.9 & 93.7 & 54.1 & +39.6 \\
Speech translation & 119 & 95.0 & 58.8 & +36.1 & 90.8 & 61.3 & +29.4 \\
\bottomrule
\end{tabular}
\end{table*}

\FloatBarrier
\section{Additional Results for MiniCPM-o}
\label{sec:appendix:minicpm}

We also report projected-logit entropy, layer-wise decision margins, and norm controls for \textbf{MiniCPM-o} on SpeechMMLU and VoiceBench BBH using the 300 aggregate items per dataset with BF16 weights and no quantization.

\subsection{Projected-Logit Entropy}
\begin{figure}[htbp]
    \centering
    \includegraphics[width=0.98\linewidth]{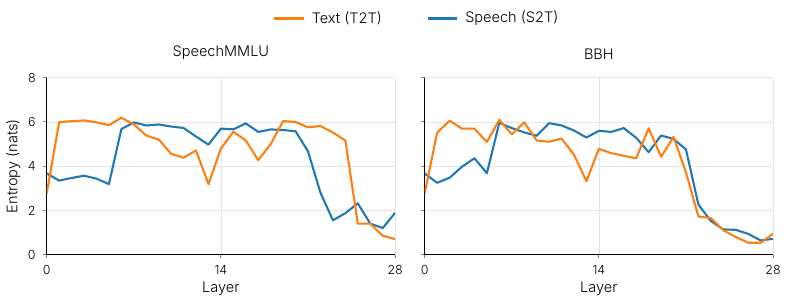}
    \caption{\textbf{MiniCPM-o layer-wise projected-logit entropy.}
    Curves show the mean over 300 items per benchmark.
    Speech and text both show a late-layer decrease, with larger differences between the routes than in Qwen2.5-Omni.}
    \label{fig:appendix:entropy_minicpm}
\end{figure}

Across both datasets, speech and text entropy both decrease in the last layers and end below 2 nats at the final coordinate.
The two curves differ more than in Qwen2.5-Omni, with mean absolute differences of 1.34 nats on SpeechMMLU and 0.85 nats on BBH.
Beyond layer 10, entropy first falls below 3 nats at layer 22 for speech and layer 25 for text on SpeechMMLU, and at layer 22 for both routes on BBH.
Option-level margins complement this distribution-wide summary by measuring preference for the correct candidate.
\Cref{fig:appendix:entropy_minicpm} provides the corresponding layer-wise entropy trajectories.

\subsection{Layer-wise Margins by Correctness Group}
\begin{figure*}[htbp]
    \centering
    \includegraphics[width=0.98\linewidth]{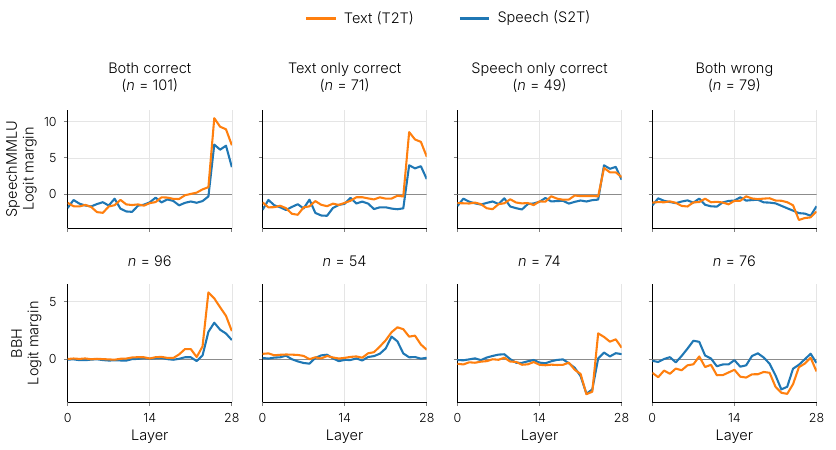}
    \caption{\textbf{MiniCPM-o layer-wise logit margin by correctness group.}
Rows are SpeechMMLU and BBH.
In the text-only-correct group the text margin exceeds the speech margin in late layers, and speech margins are smaller than text margins when both routes are correct.}
    \label{fig:appendix:margin_by_group_minicpm}
\end{figure*}

In the text-only-correct group the text margin is clearly higher than the speech margin in the late layers (mean of the last six coordinates 4.65 versus 1.56 on SpeechMMLU, and 1.92 versus 0.45 on BBH), but the speech margin stays positive on both benchmarks rather than turning negative.
Speech margins are also smaller when both routes are correct (3.67 versus 6.17, and 2.07 versus 3.82).
Between 17\% and 33\% of the answers are unparsed (\cref{tab:appendix:group_protocol}), so the groups partly reflect formatting.
\Cref{fig:appendix:margin_by_group_minicpm} shows the full groupwise margin curves.

\subsection{Pre-LN and Post-LN Norms}
\begin{figure}[htbp]
    \centering
    \includegraphics[width=0.98\linewidth]{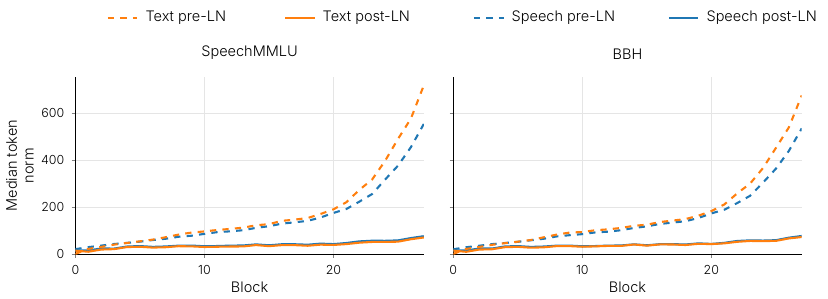}
    \caption{\textbf{MiniCPM-o pre-LN and post-LN norms.} Left is SpeechMMLU and right is BBH.
Dashed curves are the block input before the input layer norm and solid curves after it.
Post-LN norms are comparatively stable across modalities despite late pre-LN divergence.}
    \label{fig:appendix:norm_prepost_minicpm}
\end{figure}

For each item and block, we take the median L2 norm over the evidence tokens of the block input before and after the block's input layer norm, and average over the 300 items.
Pre-LN norms reach 535 to 553 for speech and 675 to 718 for text at the last block, whereas post-LN norms are 71 to 77 and differ between modalities by a factor of at most 1.6.
These curves describe activation magnitude, which does not determine the direction of the normalized states.
\Cref{fig:appendix:norm_prepost_minicpm} reports the norm-control results.

\FloatBarrier
\section{Additional Results for LLaMA-Omni}
\label{sec:appendix:llama}

We also examine \textbf{LLaMA-Omni}, using 33 hidden-state coordinates including the embedding output, with hidden dimension 4096.
We run Llama-3.1-8B-Omni in BF16 on the same 300 aggregate items with its own Llama-3 chat template and system prompt.
Its speech route always receives the 300 projected positions of a 30-second padded input, and we analyze the positions that cover audio.

\subsection{Cross-Layer CKA}
\begin{figure}[htbp]
    \centering
    \includegraphics[width=0.98\linewidth]{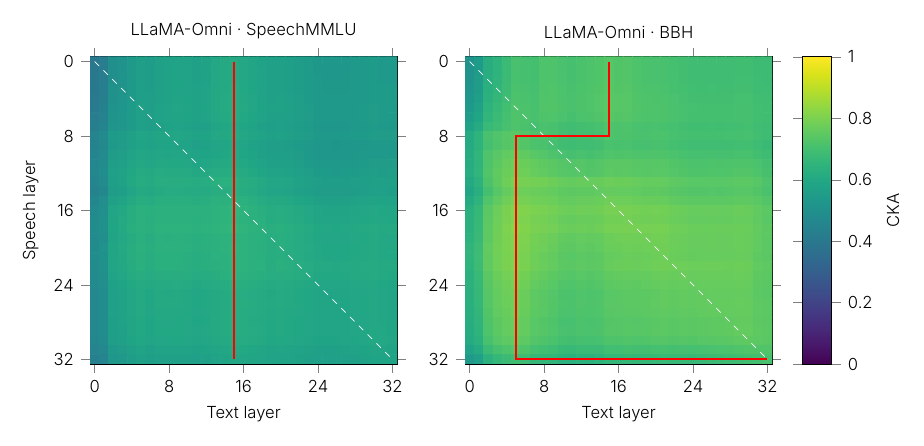}
    \caption{\textbf{LLaMA-Omni cross-layer CKA heatmaps.} Left is SpeechMMLU with DTW base layer 10 and right is BBH with base layer 20.
Red lines mark the best-matching text layer of each speech layer and dashed lines the diagonal.
Similarity is high across most layer pairs and lowest at the first two text layers.}
    \label{fig:llama_cka}
\end{figure}

Similarity is high across most layer pairs: CKA ranges from 0.40 to 0.66 on SpeechMMLU and from 0.51 to 0.82 on BBH, and only the first two text layers are clearly lower.
The best-match path maps every speech layer to text layer 15 on SpeechMMLU and to three text layers on BBH, so, compared to Qwen2.5-Omni, the high-similarity region extends across more text layers and the correspondence is less layer-specific.
\Cref{fig:llama_cka} shows the cross-layer CKA heatmaps.

\subsection{Projected-Logit Entropy}
\begin{figure}[htbp]
    \centering
    \includegraphics[width=0.98\linewidth]{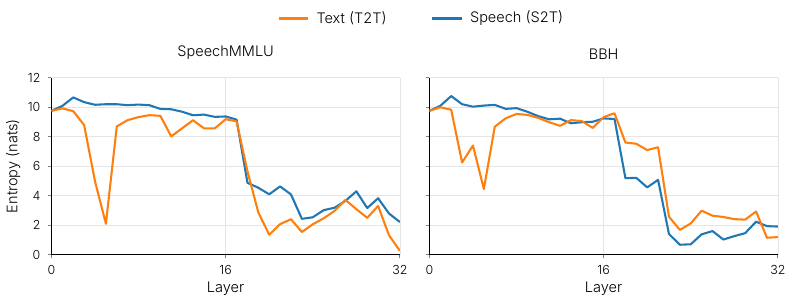}
    \caption{\textbf{LLaMA-Omni layer-wise projected-logit entropy.}
    Curves show the mean over 300 items per benchmark.
    Entropy curves are broadly comparable across modalities.}
    \label{fig:appendix:entropy_llama}
\end{figure}

Entropy trends are broadly aligned between modalities, with mean absolute differences of 1.31 nats on SpeechMMLU and 1.19 nats on BBH.
Beyond layer 10, entropy first falls below 3 nats at layer 23 for speech and layer 19 for text on SpeechMMLU, and at layer 22 for both routes on BBH.
We compare option-level margins below to examine which answer each route favors.
\Cref{fig:appendix:entropy_llama} provides the entropy curves.

\subsection{Layer-wise Margins by Correctness Group}
\begin{figure*}[htbp]
    \centering
    \includegraphics[width=0.98\linewidth]{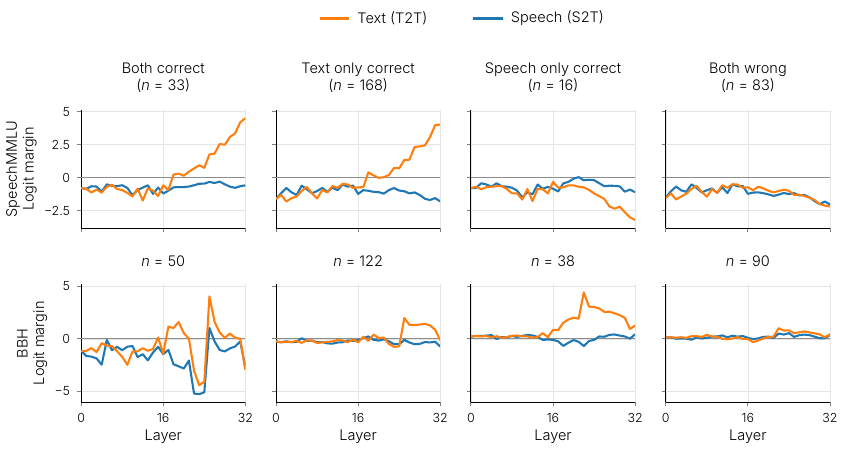}
    \caption{\textbf{LLaMA-Omni layer-wise logit margin by correctness group.}
Rows are SpeechMMLU and BBH.
The text-only-correct group exhibits negative speech margins and positive text margins in late layers.}
    \label{fig:appendix:margin_by_group_llama}
\end{figure*}

As with Qwen2.5-Omni, the text-only-correct group has a negative speech margin and a positive text margin in the late layers (mean of the last six coordinates \textminus1.50 and +3.04 on SpeechMMLU, \textminus0.44 and +1.02 on BBH).
The effect is smaller on BBH than on SpeechMMLU.
Speech margins are also negative when both routes are correct on SpeechMMLU (\textminus0.57), and 114 of the 300 speech answers on each benchmark are unparsed, so text-only-correct items dominate the groups (168 and 122).
\Cref{fig:appendix:margin_by_group_llama} shows the groupwise margin trajectories.

\subsection{Pre-LN and Post-LN Norms}
\begin{figure}[htbp]
    \centering
    \includegraphics[width=0.98\linewidth]{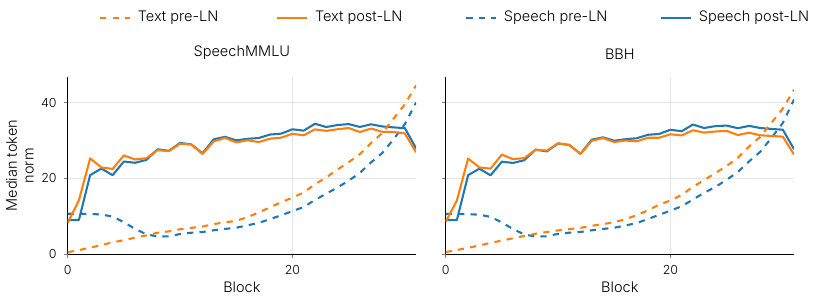}
    \caption{\textbf{LLaMA-Omni pre-LN and post-LN norms.} Left is SpeechMMLU and right is BBH.
Post-LN norms are comparatively stable across modalities while pre-LN norms grow with depth.}
    \label{fig:appendix:norm_prepost_llama}
\end{figure}

Pre-LN norms rise from about 5 at block 8 to 40 to 45 at the last block, whereas post-LN norms stay at about 20 to 35 after the first blocks.
The speech-to-text ratio of the post-LN norm ranges from 0.64 to 1.08 and ends at 1.04 to 1.05, so the cross-modal difference is small after normalization.
\Cref{fig:appendix:norm_prepost_llama} provides the norm-control visualizations.

\FloatBarrier
\section{Additional Results for Qwen2-Audio}
\label{sec:appendix:qwen2audio}

We examine \textbf{Qwen2-Audio}, whose speech accuracy on SpeechMMLU is below the 25\% uniform-choice baseline and whose modality gap is smaller on BBH.
Its CKA heatmaps show low cross-modal similarity through the middle layers, rising only in the last speech layers.
We use \emph{Representation Failure} to describe this weak-alignment pattern, in contrast to the middle-layer alignment followed by late decision instability seen in Qwen2.5-Omni and MiniCPM-o.

\subsection{Cross-Layer CKA}
\label{sec:appendix:qwen2audio_structural}

\begin{figure}[htbp]
    \centering
    \includegraphics[width=0.98\linewidth]{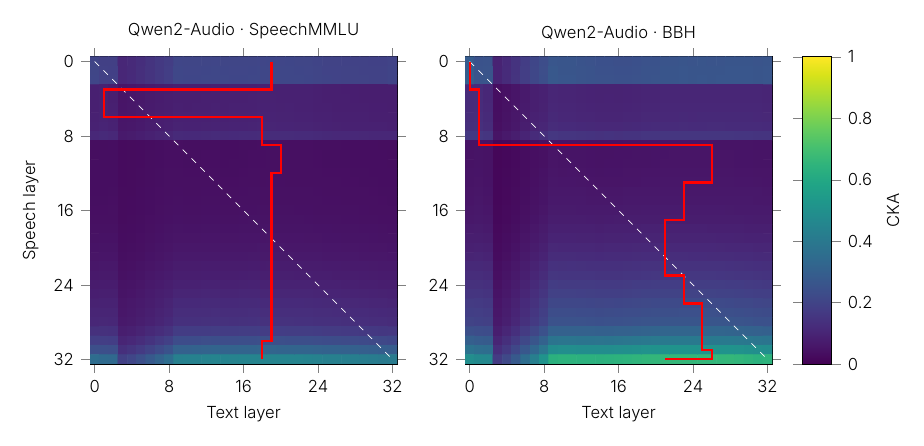}
    \caption{\textbf{Qwen2-Audio cross-layer CKA heatmaps.} Left is SpeechMMLU with DTW base layer 10 and right is BBH with base layer 20.
Red lines mark the best-matching text layer of each speech layer and dashed lines the diagonal.
Unlike Qwen2.5-Omni and MiniCPM-o, Qwen2-Audio lacks a clear high-similarity middle-layer band.}
    \label{fig:appendix:qwen2audio_cka}
\end{figure}

Unlike Qwen2.5-Omni and MiniCPM-o, Qwen2-Audio does not exhibit a clear diagonal or broad middle-layer band in its CKA heatmap.
Instead, the best-match path collapses toward a near-vertical trajectory that maps many speech layers to the same text layer.
On SpeechMMLU, 21 of the 33 speech layers share text layer 19, and speech layers 9--32 use only text layers 18--20.
On BBH, speech layers 9--32 map to text layers 21, 23, 25, and 26, as shown in \cref{fig:appendix:qwen2audio_cka}.

\paragraph{Low CKA through the middle layers.}
The key diagnostic observation is that Qwen2-Audio shows low CKA values not just in early layers, as in Phase I of the other models, but through the middle of the network: the row maxima of speech layers 10--21 stay at or below 0.07 on SpeechMMLU and 0.13 on BBH.
In Qwen2.5-Omni and MiniCPM-o, CKA heatmaps exhibit a pronounced high-similarity band in middle layers where speech and text representations converge.
Qwen2-Audio lacks this band.
Similarity rises only in the last speech layers, reaching 0.45 on SpeechMMLU and 0.66 on BBH at the final layer.
The measured speech and text trajectories therefore show weak geometric correspondence through the middle of the network.

\paragraph{Task-dependent impact.}
Weak geometric correspondence accompanies a 43.07-point gap on SpeechMMLU and a 2.2-point gap on BBH (\cref{tab:failure_modes_main}).
The task difference motivates separating geometric alignment from task-specific answer formation, as in the readouts below.

\subsection{Candidate-Only Readout}
\begin{figure}[htbp]
    \centering
    \includegraphics[width=0.98\linewidth]{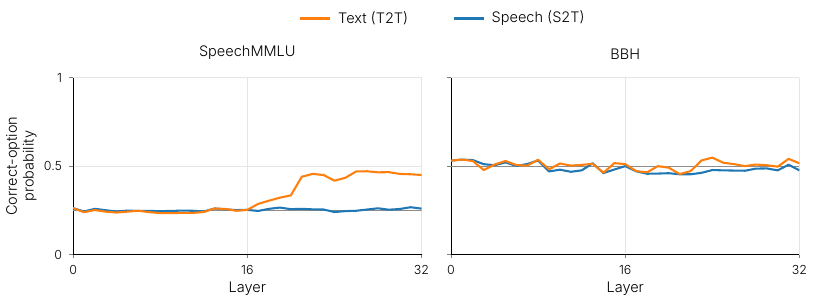}
    \caption{\textbf{Qwen2-Audio candidate-only readout.} Per-layer probability of the correct answer option among candidates for S2T and T2T at the decision position after an assistant answer prefix, averaged over 300 items per benchmark.
Thin lines mark chance, 0.25 on SpeechMMLU and 0.5 on BBH.
Speech does not raise the correct option, especially on SpeechMMLU.}
    \label{fig:appendix:qwen2audio_semantic_accessibility}
\end{figure}

On SpeechMMLU, text begins to concentrate probability mass onto the correct option in late layers, reaching 0.45 at the last layer against a chance level of 0.25, whereas speech stays within 0.24--0.27 and does not elevate the correct option above competitors.
On BBH both routes stay between 0.45 and 0.55 at every layer, around the two-option chance level of 0.5.
\Cref{fig:appendix:qwen2audio_semantic_accessibility} provides the candidate-only readout curves.

\subsection{Projected-Logit Entropy}
\begin{figure}[htbp]
    \centering
    \includegraphics[width=0.98\linewidth]{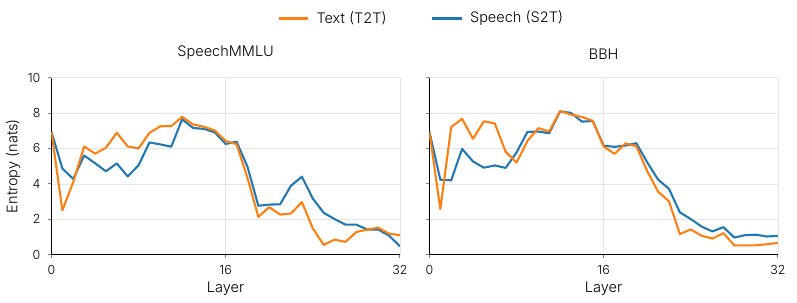}
    \caption{\textbf{Qwen2-Audio layer-wise projected-logit entropy.}
    Curves show the mean over 300 items per benchmark.
    Speech entropy is not uniformly higher than text entropy across layers.}
    \label{fig:appendix:entropy_qwen2audio}
\end{figure}

Entropy is comparable across modalities, with mean absolute differences of 0.75 nats on SpeechMMLU and 0.71 nats on BBH.
Beyond layer 10, it first falls below 3 nats at layer 19 on SpeechMMLU and at layer 23 on BBH for both routes, well before the final layers.
On SpeechMMLU, speech entropy exceeds text entropy by more than 1 nat at layers 22--26, so its late decrease is delayed, and the gap is below 1 nat from layer 27 on.
\Cref{fig:appendix:entropy_qwen2audio} shows the layer-wise entropy trajectories for Qwen2-Audio.

\subsection{Layer-wise Margins by Correctness Group}
\begin{figure*}[htbp]
    \centering
    \includegraphics[width=0.98\linewidth]{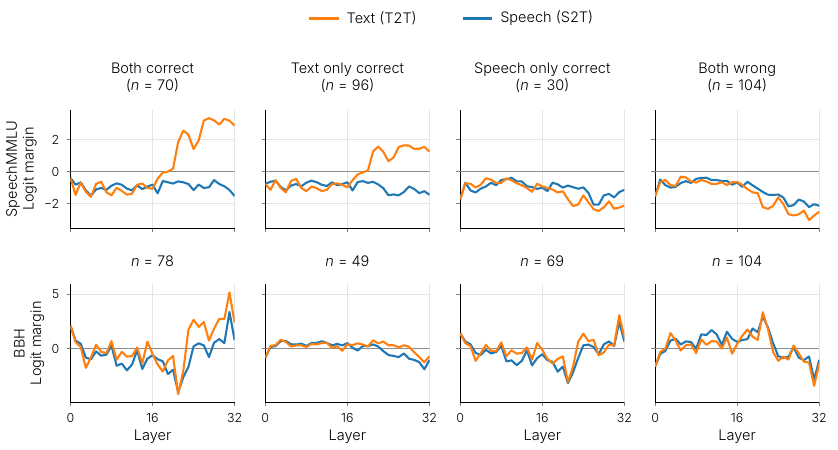}
    \caption{\textbf{Qwen2-Audio layer-wise logit margin by correctness group.}
Rows are SpeechMMLU and BBH.
Margins favor the correct option less strongly under speech than under text on SpeechMMLU.}
    \label{fig:appendix:margin_by_group_qwen2audio}
\end{figure*}

On SpeechMMLU, text margins are positive in late layers in both groups where text is correct (mean of the last six coordinates +3.15 and +1.49), while speech margins stay negative in every group (\textminus0.99 to \textminus2.03).
On BBH the margins are small, and the text margin is negative in the text-only-correct group (\textminus0.44, against \textminus1.11 for speech).
Speech-only-correct cases number 30 on SpeechMMLU and 69 on BBH, and the SpeechMMLU group shows no positive speech margin (\textminus1.50).
\Cref{fig:appendix:margin_by_group_qwen2audio} provides the groupwise margin curves.

\subsection{Pre-LN and Post-LN Norms}
\begin{figure}[htbp]
    \centering
    \includegraphics[width=0.98\linewidth]{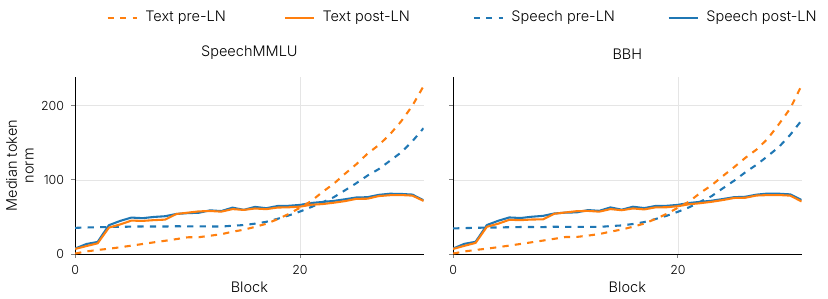}
    \caption{\textbf{Qwen2-Audio pre-LN and post-LN norms.} Left is SpeechMMLU and right is BBH.
Post-LN norms remain comparatively stable across modalities.}
    \label{fig:appendix:norm_prepost_qwen2audio}
\end{figure}

Pre-LN norms differ between modalities by up to a factor of 45 across blocks (speech-to-text ratio 0.75 to 45) and reach 170 to 180 for speech and 226 to 227 for text at the last block, whereas post-LN norms, i.e., the normalized stream entering attention, differ by at most 25\% (ratio 0.97 to 1.25).
Normalization reduces the cross-modal difference in activation magnitude.
\Cref{fig:appendix:norm_prepost_qwen2audio} shows the pre-LN and post-LN norm controls for Qwen2-Audio.

\FloatBarrier
\section{Qualitative Cases}
\label{sec:appendix:qualitative}

We inspect four examples from an earlier comparison in which the model answered correctly under text input and incorrectly under speech input.

\paragraph{Setup.}
These instances have matched prompts where T2T was correct and S2T was incorrect in that comparison, and this selection is not re-measured here.
We re-execute each item with Qwen2.5-Omni in BF16 and plot, for each layer, the probability assigned to the correct option, normalized over the answer candidates after an assistant answer prefix.
This view isolates whether the correct answer becomes a plausible competitor during intermediate computation, or whether it never gains probability mass under speech.

\begin{figure*}[htbp]
    \centering
    \includegraphics[width=0.98\linewidth]{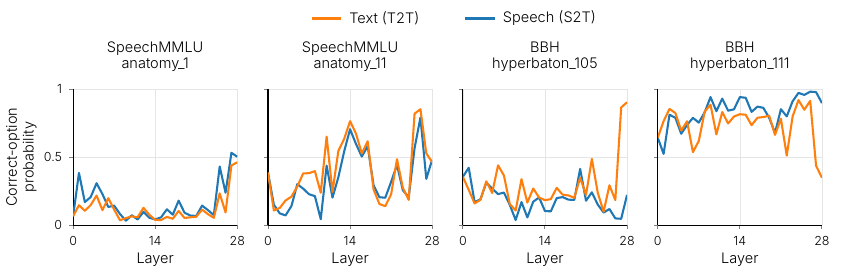}
    \caption{\textbf{Candidate-probability traces for four instances selected as text-correct and speech-wrong.}
    Curves show, for each fixed item, the probability of the correct option after normalization over answer candidates in Qwen2.5-Omni.
    Only \texttt{hyperbaton\_105} has a higher text probability at the last layer.}
    \label{fig:appendix:qualitative_case_studies}
\end{figure*}

\paragraph{Results.}
Only \texttt{hyperbaton\_105} shows a text advantage at the answer layers: the text route reaches 0.90 at the last layer while the speech route stays at 0.22.
In \texttt{hyperbaton\_111} the order reverses, with the speech route at 0.90 and the text route at 0.35 at the last layer.
In the two SpeechMMLU cases the routes end at similar probabilities after fluctuating through the middle and late layers (speech and text 0.50 and 0.46 for \texttt{anatomy\_1}, 0.48 and 0.46 for \texttt{anatomy\_11}).
This readout is a candidate probability at a single position and does not reproduce the free-generation outcome that selected these items, so the traces show layer-to-layer variability rather than a general speech failure.
\Cref{tab:appendix:qualitative_cases,fig:appendix:qualitative_case_studies} summarize the selected prompts and the corresponding per-layer traces.
In \texttt{web\_of\_lies\_74}, baseline S2T answers ``No'' while gold-text T2T answers ``Yes'' in the 1,024-token diagnostic, and tempo 1.2\texttimes{} flips S2T to ``Yes'' without changing lexical content.
Under the shared-instruction evaluation, both routes answer this item correctly.

\begin{table}[htbp]
\centering
\small
\setlength{\tabcolsep}{3pt}
\renewcommand{\arraystretch}{1.15}
\caption{\textbf{Qualitative cases (text correct, speech wrong).} We summarize selected questions where T2T answered correctly but S2T failed in the earlier comparison that selected them.
The accompanying figure plots the per-layer probability of the correct option, normalized over answer candidates.}
\vspace{2mm}
\begin{tabular}{llp{0.56\linewidth}}
\toprule
\textbf{Dataset} & \textbf{Item} & \textbf{Question summary} \\
\midrule
SpeechMMLU & \texttt{anatomy\_1} & Which craniofacial condition is associated with a ``dished face'' profile? \\
SpeechMMLU & \texttt{anatomy\_11} & Which signs follow damage to the third cranial nerve after an orbital injury? \\
BBH & \texttt{bbh\_hyperbaton\_105} & Choose the correct ordering of adjectives describing a repulsive, normal-size, green, old-fashioned cardboard surfboard. \\
BBH & \texttt{bbh\_hyperbaton\_111} & Choose between ``little black driving ship'' and ``driving little black ship.'' \\
BBH & \texttt{bbh\_web\_of\_lies\_74} & Jamey tells the truth. Raymond says Jamey lies. Audrie says Raymond lies. Michaela says Audrie lies. Rashida says Michaela lies. Does Rashida tell the truth? \\
\bottomrule
\end{tabular}
\label{tab:appendix:qualitative_cases}
\end{table}

\FloatBarrier
\section{Implementation Details}
\label{sec:appendix:implementation}

\subsection{Hyperparameters and Setup}
\label{sec:appendix:setup_details}
Key implementation choices that affect all analyses are reported below.
\Cref{tab:hyperparams} lists the shared settings used for DTW and CKA analyses in this paper.

\begin{table}[htbp]
\centering
\caption{\textbf{Summary of implementation hyperparameters.}}
\label{tab:hyperparams}
\vspace{1mm}
\small
\setlength{\tabcolsep}{6pt}
\renewcommand{\arraystretch}{1.15}
\begin{tabular}{lp{0.55\linewidth}}
\toprule
\textbf{Parameter} & \textbf{Value or Method} \\
\midrule
\multicolumn{2}{l}{\textit{Datasets and Audio}} \\
SpeechMMLU & Dataset-provided paired \texttt{question\_text}, answer options, \texttt{question\_audio}, and speaker identifier from MiMo-Audio-Eval \\
VoiceBench BBH & Dataset-provided paired \texttt{prompt}, reference answer, and audio clip for \texttt{hyperbaton}, \texttt{navigate}, \texttt{sports\_understanding}, and \texttt{web\_of\_lies} \\
T2T Prompt & The paired text field corresponding to the audio item. No ASR transcript is used for gold-text T2T \\
\midrule
\multicolumn{2}{l}{\textit{Hidden State Extraction}} \\
Extraction Window & Last user segment, i.e., question turn only \\
Valid Tokens & Positions where \texttt{attention\_mask=1} \\
\midrule
\multicolumn{2}{l}{\textit{DTW Alignment}} \\
Base Layer $\ell_0$ & Layer 10 for SpeechMMLU, Layer 20 for BBH \\
Metric & Cosine similarity with per-row normalization \\
Step Penalty $\lambda$ & 0, i.e., standard DTW \\
Band Constraint & None, i.e., unconstrained \\
Many-to-one Aggregation & Speech frames aligned to the same text token are mean-pooled before CKA, standardized L2, and norm computation \\
\midrule
\multicolumn{2}{l}{\textit{Cross-Layer Aggregation}} \\
Item Selection & Fixed set of 100 items with seed 0 \\
CKA Method & Per-example linear CKA with double-centering, averaged across the selected items \\
\midrule
\multicolumn{2}{l}{\textit{ASR-to-LLM Control}} \\
Transcript Prompting & Same downstream LLM prompted with ASR transcript. Explicit question/option markers are preserved or restored when present \\
\bottomrule
\end{tabular}
\end{table}

\subsection{Compute Resources}
\label{sec:appendix:compute}

The confirmatory alignment, paired-readout, and intervention runs used NVIDIA A800 GPUs with 80 GB memory, with independent conditions or shards distributed over four GPUs.
Each native-center alignment cell used one GPU and 100 items.
Its recorded processing interval was 178 seconds for Qwen2.5-Omni on BBH, 342 seconds on SpeechMMLU, 175 seconds for MiniCPM-o on BBH, and 327 seconds on SpeechMMLU.
These intervals run from metadata initialization before model loading through completion.
The four cells therefore used about 0.284 GPU-hours within those intervals.
The study also includes behavioral inference, other diagnostics, and discarded pilots, for which a complete total compute accounting is unavailable.

\FloatBarrier
\section{Broader Impact}
\label{sec:appendix:broader_impact}

Separating internal similarity from answer reliability can help developers evaluate speech interfaces before using them for consequential tasks.
Our results also show that a change in an internal diagnostic need not improve accuracy, which can prevent misplaced confidence in a repair.
Suppressing acoustic variation indiscriminately could remove information needed for emotion, intent, or accessibility-related communication.
The benchmark setting does not measure performance disparities across accents or disabilities, so the reported diagnostics should not be treated as evidence of equitable deployment performance.

\FloatBarrier
\section{Existing Assets and Licenses}
\label{sec:appendix:assets}

We use existing pretrained models and benchmark items without training new model weights or collecting new human-subject data.
\Cref{tab:appendix:assets} records the licenses stated by the linked providers, checked in October 2026.
The MiMo-Audio-Eval toolkit license does not by itself specify the license of SpeechMMLU's synthesized audio.
OpenAI's original Whisper release states MIT for code and weights, while the Hugging Face model card labels the checkpoint Apache-2.0.

\begin{table}[htbp]
\centering
\small
\caption{\textbf{Provider-stated licenses of existing assets.}
Links point to the official model cards, dataset cards, or repositories.
These are source declarations rather than a complete audit of every underlying asset.}
\label{tab:appendix:assets}
\begin{tabular*}{\linewidth}{@{\extracolsep{\fill}}lp{0.56\linewidth}@{}}
\toprule
Asset and source & Stated license or condition \\
\midrule
\href{https://huggingface.co/Qwen/Qwen2.5-Omni-7B}{Qwen2.5-Omni-7B} & Apache-2.0 \\
\href{https://huggingface.co/openbmb/MiniCPM-o-2_6}{MiniCPM-o 2.6} & Apache-2.0 for code and weights \\
\href{https://huggingface.co/Qwen/Qwen2-Audio-7B-Instruct}{Qwen2-Audio-7B-Instruct} & Apache-2.0 \\
\href{https://huggingface.co/ICTNLP/Llama-3.1-8B-Omni}{LLaMA-Omni} & Code: Apache-2.0. Model: academic, noncommercial use with citation \\
\href{https://github.com/openai/whisper}{Whisper-large-v2} & MIT in the OpenAI release, with differing \href{https://huggingface.co/openai/whisper-large-v2}{checkpoint-card metadata} \\
\href{https://github.com/XiaomiMiMo/MiMo-Audio-Eval}{MiMo-Audio-Eval toolkit} & Apache-2.0 \\
\href{https://huggingface.co/datasets/XiaomiMiMo/SpeechMMLU}{SpeechMMLU} & Audio license not stated in the dataset card \\
\href{https://huggingface.co/datasets/cais/mmlu}{MMLU} & MIT \\
\href{https://huggingface.co/datasets/hlt-lab/voicebench}{VoiceBench} / \href{https://github.com/suzgunmirac/BIG-Bench-Hard}{BBH} & Apache-2.0 / MIT \\
\href{https://huggingface.co/datasets/ddwang2000/MMSU}{MMSU} & MIT \\
\bottomrule
\end{tabular*}
\end{table}

\end{document}